\documentclass[journal]{IEEEtran}
\usepackage[T1]{fontenc}
\usepackage{algorithm}
\usepackage{algorithmic}
\usepackage{hyperref}
\usepackage{subfigure}
\usepackage{multirow}
\usepackage{makecell}
\usepackage{amsmath,amstext}
\usepackage{amssymb}
\usepackage{tabularx}
\usepackage{graphicx}
\usepackage{enumitem}
\hypersetup{
    colorlinks=true,
    urlcolor=blue
}

\begin{document}
\bstctlcite{IEEEexample:BSTcontrol}

\title{Large-Language-Models-as-a-Judge in Theory-Agnostic Adaptive Metric-Alignment for Prototypical Networks in Personality Recognition}

\author{
    \IEEEauthorblockN{
        Jing Jie Tan\IEEEauthorrefmark{1}, 
        Ban-Hoe Kwan\IEEEauthorrefmark{1}, 
        Danny Wee-Kiat Ng\IEEEauthorrefmark{1}, 
        Yan-Chai Hum\IEEEauthorrefmark{1},\\
        Shih-Yu Lo \IEEEauthorrefmark{2}, 
        Po-An Chen \IEEEauthorrefmark{3},
        Noriyuki Kawarazaki\IEEEauthorrefmark{4}, 
        Kosuke Takano\IEEEauthorrefmark{4},
        Anissa Mokraoui\IEEEauthorrefmark{5}
    }
    
    \IEEEauthorblockA{\IEEEauthorrefmark{1}Department of Mechatronics and Biomedical Engineering, Lee Kong Chian Faculty of Engineering and Science,\\ 
    Universiti Tunku Abdul Rahman, Malaysia}
    
    \IEEEauthorblockA{\IEEEauthorrefmark{2}Institute of Communication Studies, National Yang Ming Chiao Tung University, Taiwan }
    
    \IEEEauthorblockA{\IEEEauthorrefmark{3}Institute of Information Management, National Yang Ming Chiao Tung University, Taiwan }

    \IEEEauthorblockA{\IEEEauthorrefmark{4}Faculty of Information Technology, Kanagawa Institute of Technology, Japan}
    
    \IEEEauthorblockA{\IEEEauthorrefmark{5}Laboratoire de Traitement et Transport de l'Information, Université Sorbonne Paris Nord, France}
    
    Email: tanjingjie@1utar.my, \{kwanbh, ngwk, humyc\}@utar.edu.my,\{shihyulo, poanchen\}@nycu.edu.tw,\{kawara@rm, takano@ic\}.kanagawa-it.ac.jp, anissa.mokraoui@univ-paris13.fr
}

\maketitle

\begin{abstract}
Personality recognition has traditionally been constrained by theory-dependent formulations, where models are trained to fit predefined psychological taxonomies rather than uncovering shared underlying behavioral structure. This limits generalization, as personality itself is better understood as theory-invariant, emerging from stable psychological patterns that should manifest consistently across different frameworks, while existing annotations reflect only partial and sometimes inconsistent views of the same latent traits. In this work, we introduce \textbf{JAM} ((J)udge for (A)daptive (M)etric-Alignment), a theory-agnostic framework that shifts learning from adapting to predefined personality theories toward discovering unified latent ``pseudo-facets'' that capture shared psychological structure. Rather than constraining the model to any personality taxonomy during training or inference, the framework learns generalizable psychological representations and can infer an individual's latent psychological profile directly from the textual samples, without requiring theory-specific labels. JAM achieves this through an Attention-Pooled Graph Prototypical Network that learns structured representations via clustering in embedding space, together with a Cross-Theory Harmonization (CTH) approach that integrates (i) Human-Guided Linkage and (ii) Machine-Induced Consensus to unify heterogeneous datasets without relying on predefined labels. To further improve robustness and data quality, we incorporate an LLM-as-a-Judge mechanism operating in two configurations, (i) LLM-before-the-loop and (ii) LLM-in-the-loop which identifies ambiguous, mislabeled, and boundary samples to guide adaptive metric learning. Experiments on Essays and Kaggle personality datasets show that JAM improves cross-framework generalization and performance, establishing a strong step toward theory-agnostic personality inference and supporting low-resource personality theories. The related code repository, model weights, and artifacts are available at \url{https://research.jingjietan.com/JAM}.
\end{abstract}


\section{Introduction}
Personality recognition has become increasingly important, especially in recommendation systems \cite{dhelim}. By understanding user personalities, these systems can provide personalized suggestions, enhancing user satisfaction and trust. Understanding user personality is crucial for delivering a superior user experience, making this an important area of study \cite{aitbaha}. By tailoring interactions and recommendations to individual personality traits, AI systems and robots can achieve higher levels of personalization, leading to increased user satisfaction and trust \cite{li2024exploringpersonalitydrivenpersonalizationxai}. Furthermore, incorporating personality traits into recommendation algorithms can help address issues such as the cold start and data sparsity problems, resulting in more accurate and user-centric recommendations \cite{Omidvar,dhelim2021bigfivemptieysenckhexaco}. Additionally, this approach can contribute to explainable AI by providing clearer explanations for the recommendations made.

Despite these advancements, current personality-recognition models remain constrained by several fundamental limitations. Most approaches are built around specific psychological theories, such as the Big-5 or MBTI, limiting their ability to generalize across datasets and cultural contexts. This dependence is further exacerbated by the scarcity of large-scale annotated data. The widely used\textbf{ myPersonality dataset} \cite{stillwell2015mypersonality}, which originally contained data from millions of users, was \textbf{discontinued} due to privacy concerns, leaving only relatively small public datasets. Moreover, existing methods treat personality recognition as a static classification task, without leveraging psychological insights, limiting their generalizability.

These limitations motivate a \textbf{theory-agnostic framework} that eliminates dependence on predefined personality taxonomies during both training and inference. Instead of learning theory-specific representations, the model discovers latent \textit{pseudo-facets} that capture underlying psychological structure across heterogeneous data sources. At inference time, the model directly infers an individual's latent psychological profile from behavioral or textual samples without requiring theory-specific labels or adherence to a particular personality framework. This design enables the integration of heterogeneous datasets while improving generalization across existing personality theories.

\subsection{Contributions}
In this study, we make the following key contributions:
\begin{enumerate}
    \item We incorporate a psychology-based methodology for integrating datasets constructed under different personality theories and generalize the model through human-guided linkage, supporting low-resource personality theory.
    
    \item We enhance the prototypical network by incorporating machine-induced consensus for pseudo-facet construction, improving cross-theory harmonization and mitigating class imbalance, leading to improved performance.

    \item We investigate the LLM-as-a-Judge mechanism, assessing its comparative effectiveness when applied during the pre-learning stage (LLM-before-the-loop) versus within the learning process (LLM-in-the-loop) for filtering ambiguous or noisy personality-related text samples.
\end{enumerate}

\section{Literature Review}

\subsection{Personality Theory} \label{sec:pt}
Recent studies have demonstrated that humans express their personality through language use, providing a rich source of data for personality recognition \cite{Phan,Dalvi,lampropoulos2022impact,Ahmed}. The rise of social media platforms has provided a rich source of data to understand how particular individuals often leave behind a personality footprint through their online activities \cite{Phan}. This creates a significant opportunity to extract personality information from such indirect background data, which is particularly beneficial for initializing personality-aware functionalities in new robots or applications. There are two primary types of assessments to measure, quantify, and classify personality traits \cite{spielman2021personality,robins2023kinds}: 
\begin{itemize}
    \item Self-report tests: These tests require test-takers to understand the provided statements and evaluate how well they describe themselves. The results can be quantitatively standardized, ensuring high reliability and validity \cite{Waugh}.
    \item Projective tests: These tests involve asking test-takers to provide their interpretations of scenes, scenarios, or objects \cite{mukhtasar2021study}. This type of test considers various aspects, such as tone, message, or body language, making it more capable of addressing potential issues like misinterpretation of questions or dishonesty from the test-taker \cite{Tett}.
\end{itemize}

\begin{figure}[ht]
    \centering
    \includegraphics[width=1\linewidth]{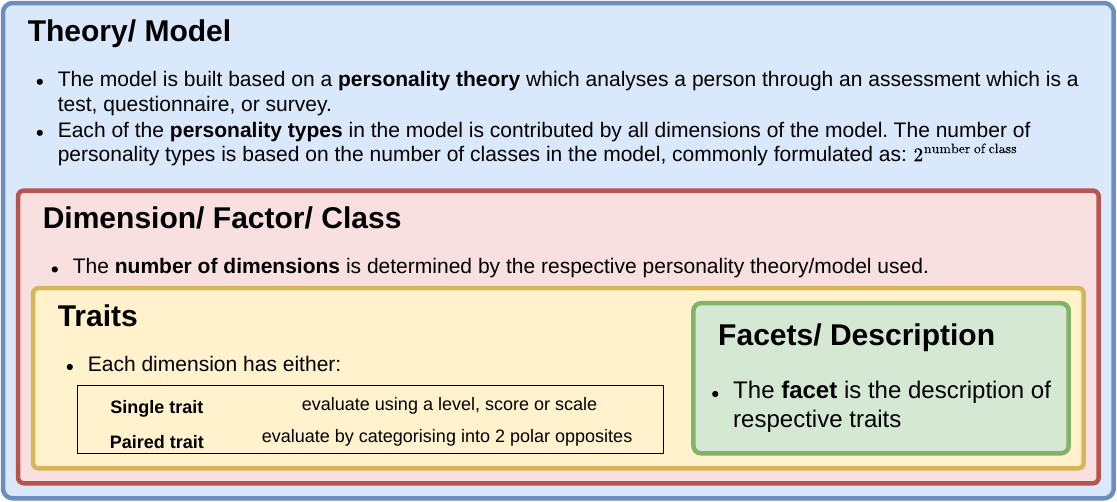}
    \caption{Overview of the relationship, terminology, and theoretical structure of personality assessment models, illustrating how theories define dimensions, traits, facets, and personality types.}
    \label{fig:litpr}
    \end{figure}

Various personality theories have been proposed to model personality using distinct sets of dimensions \cite{Roberts, cervone2022personality}. For instance, the Myers–Briggs Type Indicator (MBTI), sometimes described in relation to 4 dichotomies, categorizes personality across paired preferences \cite{Myers1962, Furnham2017}. The Big-5 framework (OCEAN) defines five broad traits: Openness to Experience, Conscientiousness, Extraversion, Agreeableness, and Neuroticism \cite{Goldberg1993, LUO2023106968}. The HEXACO model extends this structure by adding Honesty–Humility to Emotionality, Extraversion, Agreeableness, Conscientiousness, and Openness \cite{Ashton2004, william}. Table \ref{tab:personality_mapping} summarizes the correspondence among these three major frameworks, offering a general comparison of their dimensions. Basically, facets are commonly regarded as the lower-level constructs or descriptions that collectively define a broader personality dimension or trait. They provide a more granular representation of personality by capturing specific behavioral and psychological tendencies within each dimension. The terminology can be summarized as in Fig. \ref{fig:litpr}.

\begin{table*}
    \small
        \caption{Mapping the dimensions of three prominent personality frameworks: Big-5, MBTI, and HEXACO.}
    \centering
    \begin{tabularx}{\linewidth}{|X|l|l|l|l|}
    \hline
    \textbf{Symbol} & \textbf{Big-5}& \textbf{MBTI}      & \textbf{HEXACO}  & \textbf{General Description}\\ \hline
    O& Openness to Experience (O)  & Sensors (S)-Intuitive (N) & Openness (O)     & Creativity, curiosity, novelty    \\ \hline
    C& Conscientiousness (C)& Perceivers (P)-Judgers (J)& Conscientiousness (C)   & Organise, responsibility, discipline   \\ \hline
    E& Extraversion (E)    & Introverts (I)-Extroverts (E)    & Extraversion (X) & Sociability, energy, assertiveness\\ \hline
    A& Agreeableness (A)   & Thinkers (T)-Feelers (F)  & Agreeableness (A)& Empathy, kindness, cooperativeness\\ \hline
    N& Neuroticism (N)     & -    & Emotionality (E) & Sensitivity, anxiety, hostility  \\ \hline
    H& -     & -    & Honesty-Humility (H)    & Sincerity, fairness, modesty      \\ \hline
    \end{tabularx}
    \begin{minipage}{\linewidth}
\vspace{1em} 
\scriptsize
\raggedright
  Note: The correlations presented are intended as reference points, aligning these theories to the widely recognized Big-5 framework. This alignment supports evaluating model performance across diverse personality theories, fostering comparative analysis and deeper insights into their unique characteristics.
\end{minipage}
  
    \label{tab:personality_mapping}
\end{table*}

These mappings should be interpreted cautiously: they reflect partial conceptual overlap rather than equivalence. For example, Big-5 Openness includes facets such as curiosity and preference for novelty, whereas MBTI's Sensing–Intuition dimension distinguishes concrete, detail-oriented perception from abstract, future-oriented thinking. Similarly, related constructs are distributed differently across frameworks; emotional reactivity in Big-5 Neuroticism only loosely corresponds to the MBTI Thinking–Feeling dimension, which primarily reflects decision-making style rather than affective stability.

This highlights a \textbf{limitation}: \textbf{personality models are human-constructed abstractions} rather than directly observable ground truth categories. Consequently, their use in machine learning can introduce \textbf{systematic noise and inconsistency}, especially at the facet level where definitions differ across studies. In practice, available datasets seldom provide sufficiently fine-grained and consistently annotated signals to support reliable bottom-up learning of stable psychological components, not only due to annotation cost but also because personality is inherently composite and context-dependent. To our knowledge, no prior work has directly addressed this mismatch between psychologically defined trait hierarchies and their operationalization in data-driven settings, or proposed effective strategies to mitigate the resulting label ambiguity.

\subsection{Related Algorithms}
Personality recognition has evolved alongside advances in natural language processing and representation learning \cite{Tan2025psychonlu}. Early research primarily focused on developing task-specific models for predicting personality labels defined by a particular framework, such as the Big-5 or MBTI. As language models became increasingly capable of capturing semantic and contextual information, researchers began adopting pretrained Transformer architectures as general-purpose representations for personality-related tasks. More recently, large language models have enabled prompting-based inference, reducing reliance on framework-specific training procedures and expanding the possibility of transferring personality knowledge across different theoretical formulations.

To contextualize these developments, this section first reviews the language-model foundations underlying modern personality recognition, including (i) sentence transformer language models (encoder-only models) and (ii) autoregressive transformer large language models (decoder-only models). It then surveys (iii) prior personality recognition methods built upon these representations, followed by (iv) recent generative approaches that increasingly support flexible personality inference. Together, these developments provide the foundation for investigating personality recognition beyond a single predefined personality theory.

\subsubsection{Sentence Transformer Language Model (Encoder-only Model)}
These models are trained on large corpora of text data, enabling them to capture complex linguistic patterns and relationships \cite{MiniLMNEURIPS2020_3f5ee243,MPNetNEURIPS2020_c3a690be,Sentencet5,rbear-cook-2023-fine,r10.3390/math12243990}. The produced embeddings represent the semantic meaning of the input text, facilitating downstream tasks such as text classification, clustering, and similarity analysis. \textbf{MiniLM} \cite{MiniLMNEURIPS2020_3f5ee243} compresses large Transformer models through deep self-attention distillation, where a small student model mimics the self-attention of the teacher's last layer. It retains over 99\% accuracy on SQuAD 2.0 and GLUE benchmarks while reducing Transformer parameters and computation by 50\%. Moreover, \textbf{MPNet} \cite{MPNetNEURIPS2020_c3a690be} integrates permuted language modeling (PLM) and auxiliary position information, improving dependency modeling and reducing position discrepancy. It surpasses BERT, XLNet, and RoBERTa across GLUE and SQuAD. Furthermore, \textbf{Sentence-T5} \cite{Sentencet5} explores sentence embeddings from T5 models, introducing an encoder-only approach that outperforms Sentence-BERT and SimCSE in STS tasks. Scaling T5 to billions of parameters further enhances performance, setting new state-of-the-art results for sentence embeddings. Given the efficiency and strong semantic representation capabilities of these models, we select a relatively lightweight yet effective encoder-only model tailored for the personality domain, ensuring optimal performance while balancing computational efficiency.

\subsubsection{Autoregressive Transformer Large Language Model (Decoder-only Model)}

Generative Pretrained Transformer 3 (GPT-3) was the first to demonstrate effectiveness in Natural Language Understanding (NLU) through its few-shot learning capability, facilitated by techniques such as Chain-of-Thought (CoT) prompting. This family of decoder-only models enhances reasoning, coherence, and adaptability in tasks such as emotion detection, sentiment analysis, and personality recognition \cite{app12178662, sent10623654, li2024eerpdleveragingemotionemotion}. The open-source LLaMA model introduced a family of models with varying parameter sizes, achieving competitive performance while maintaining efficiency in training and inference \cite{llama}. Following this, Mistral AI released the Mistral 7B model, which, despite its relatively smaller size, demonstrated remarkable capabilities in code generation, mathematics, and reasoning tasks, benefiting from an extended context window of 128k tokens \cite{mistral}. Qwen models have been proposed as strong open-weight alternatives, exhibiting robust performance across reasoning, multilingual understanding tasks \cite{qwenBai2023}. OpenAI's GPT-4 further pushed the boundaries by introducing multimodal capabilities, allowing the model to process both text and images, thereby enhancing its applicability across diverse domains \cite{gpt4}.

Later, reasoning models further extended their capabilities in scientific and mathematical contributions. OpenAI's o1 model introduced a novel approach by generating extended chains of thought before arriving at a final answer, thereby improving the model's reasoning depth and accuracy \cite{o1}. DeepSeek's R1 model emerged as a notable open-source contribution, achieving performance comparable to OpenAI's o1 model across mathematics, coding, and reasoning tasks \cite{deepseekr1} with a relatively low computational complexity. The R1 model employs a unique training methodology that emphasizes reinforcement learning to enhance its reasoning capabilities. These developments underscore a growing trend toward integrating advanced reasoning processes within large language models, aiming to improve their problem-solving abilities and decision-making processes.

\subsubsection{Prior Works for Personality Recognition}

Prior research has consistently highlighted the importance of large, high-quality datasets for training machine learning models that generalize effectively \cite{Tan2026cross,Tanfocal2025}. However, in privacy-sensitive domains, the development and utilization of such datasets are often constrained by regulatory and ethical considerations. In contrast, other research areas are better positioned to achieve scale by aggregating or integrating existing public datasets. A frequently cited example is the \textbf{myPersonality Project} \cite{stillwell2015mypersonality}, once a prominent open dataset for computational personality research, which was discontinued in 2018 due to increasing challenges related to regulatory compliance, controlled data access, and governance obligations. In personality recognition, we are limited by personality theory; hence, few researchers are tackling algorithmic improvements, including architecture, feature exploration, graph neural networks, loss reweighted contributions, etc.

Building on the importance of data and representation, prior work has explored a range of modeling approaches for personality classification from text. Early advances include BiLSTM-based models \cite{Khattak2023}, which leverage bidirectional context to outperform traditional machine learning methods such as SVM, RF, and DT, highlighting the role of sequential modeling and hyperparameter tuning. To further enhance representation learning, \cite{Personality2vec} proposes Personality2vec, which integrates semantic, linguistic, and structural information from social networks through biased random walks and skip-gram modeling, demonstrating robustness particularly in data-scarce settings. After that, \cite{Tanfocal2025} further explored the dataset splitting and proposed a stratification feature, as well as focal loss for reweighting the contribution in the neural network, and analyzed the evaluation metrices.

Subsequent studies have focused on enriching deep contextual representations with complementary linguistic and structural information. For example, a hybrid Transformer–BLSTM framework \cite{kerz-etal-2022-pushing} integrates psycholinguistic features with attention mechanisms to improve both predictive performance and interpretability. Graph-based methods have also emerged as an effective direction. KGrAt-Net \cite{Ramezani2022} leverages knowledge graph attention over DBpedia entities to model richer semantic relationships, while TranSentGAT \cite{TranSentGAT} combines BERT with sentiment knowledge and graph attention to enhance contextual representations and improve personality prediction.

Nonetheless, these advances in modeling and feature integration \textbf{do not resolve the core bottleneck of limited and fragmented data}. Performance in personality recognition remains constrained by dataset scale, diversity, and ecological validity under strict privacy and governance requirements. From this perspective, expanding datasets in a cost-efficient and flexible manner is essential to overcoming current theoretical and empirical limits. However, collecting such data is often difficult, particularly for low-resource settings, due to time, cost, and regulatory constraints, while cross-institutional approaches such as federated learning remain limited in practice (to the best of our knowledge), partly because of heterogeneous theoretical frameworks and incompatible data assumptions. These challenges motivate a shift toward more \textbf{theory-agnostic} approaches.

\subsubsection{Generative Approaches to Personality Recognition}

Recent work on generative AI for personality inference has explored both data-centric and architecture-centric improvements. Early studies focus on data augmentation and heterogeneous graph-based models for dialogue understanding. For instance, Wu et al.~\cite{Wu2024} introduce personality trait interpolation to generate synthetic training data and propose HC-GNN to capture both inter- and intra-speaker dependencies, improving conversational personality recognition. Similarly, Semi-PerGCN~\cite{Zhu_Xia_Li_Zhang_Wu_2024} adopts a semi-supervised framework with multi-view graph augmentation and heterogeneous graph construction, addressing limited labeled data and enhancing robustness in low-resource settings. Collectively, these approaches reflect a transition from sequential modeling toward graph-based and hybrid architectures, with increasing emphasis on representation richness and data efficiency.

More recent studies have investigated large language models as direct inference engines for personality recognition. ChatGPT has been shown to exhibit strong zero-shot capability and partial interpretability in personality prediction tasks~\cite{ji2023chatgptgoodpersonalityrecognizer}. Related work integrates emotional knowledge with structured prompting strategies for trait inference~\cite{li2024eerpdleveragingemotionemotion}, while PICEPR~\cite{Tan2025picepr} not only proposes embedding-based knowledge elicitation, but it also demonstrates that modular prompting can improve classification performance. 

Compared with traditional approaches that require task-specific feature engineering or model training, these methods leverage in-context learning, enabling models to infer personality labels directly from instructions, label descriptions, or a small number of exemplars. Consequently, prompting-based approaches move toward a more theory-agnostic paradigm, as the same pretrained model can be adapted to different personality frameworks by modifying the prompt and label definitions rather than redesigning or retraining the prediction architecture. While the provided labels may still originate from a specific theory, the underlying inference mechanism is not inherently bound to a specific personality theory. Through in-context learning, the same model can perform personality inference in few-shot settings using a small number of labeled examples or in zero-shot settings using only trait descriptions and task instructions, without requiring model retraining.

However, these approaches also introduce important limitations. First, decoder-only large language models rely on \textbf{inference-time reasoning}, which is computationally expensive and often impractical for large-scale deployment. Second, they do not produce \textbf{explicit intermediate representations} of psychological states, limiting their usefulness for downstream modeling of structured cognitive or behavioral processes. More critically, since these models are trained on internet-scale corpora, there is a non-trivial risk of implicit exposure to similar evaluation data, weakening the assumption of strict zero-shot generalization and introducing \textbf{potential dataset leakage} concerns. These issues suggest that encoder-based models remain necessary for stable and controllable representation learning in personality recognition systems.

\section{Methodologies}

\begin{figure*}[ht]
    \centering
    \includegraphics[width=\linewidth]{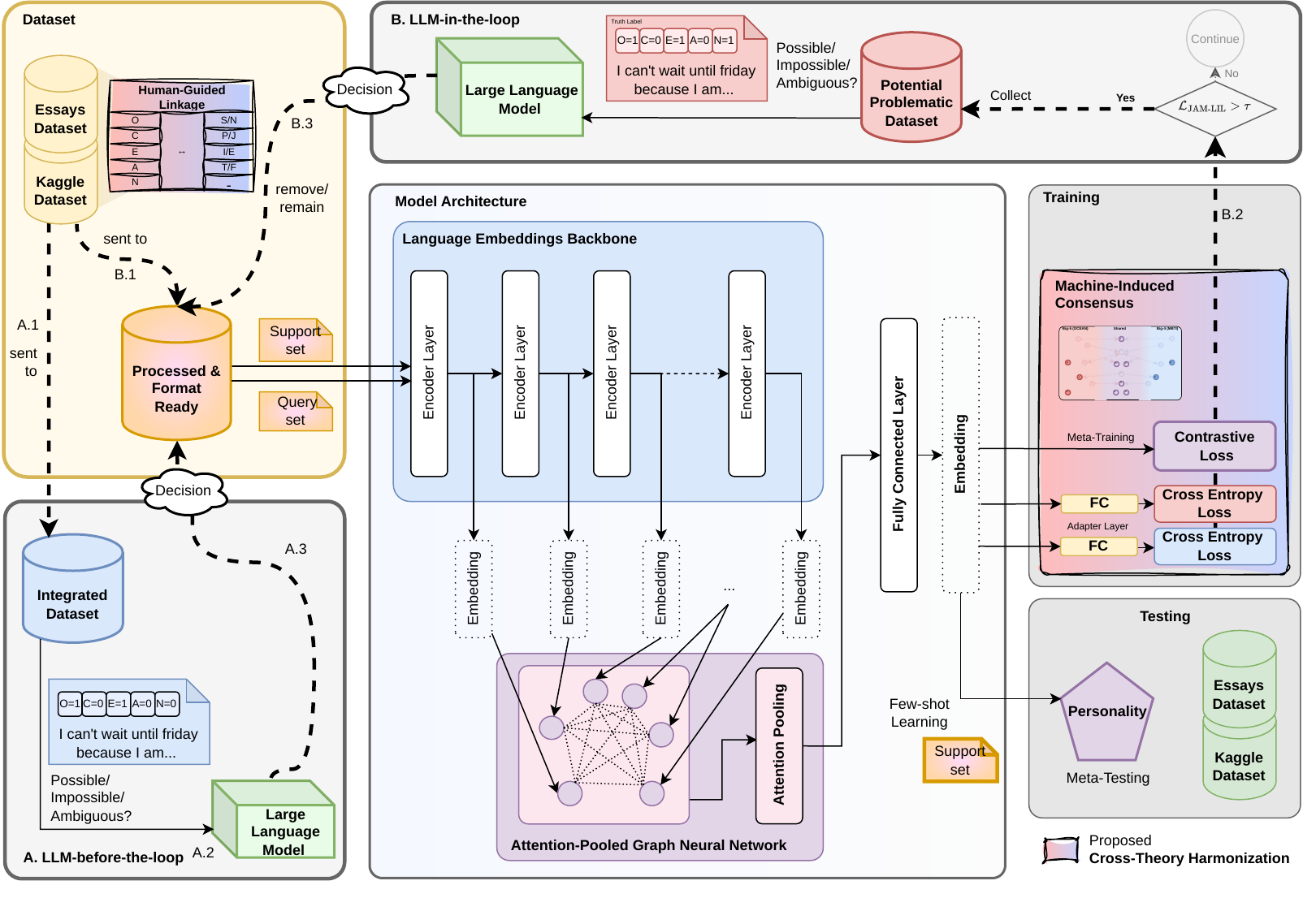}
    \caption{Overview of the proposed JAM architecture for theory-agnostic personality recognition. The framework integrates a language embedding backbone with an attention-pooled graph prototypical network to learn representations that capture latent pseudo-facets through clustering in the embedding space. A Cross-Theory Harmonization module is introduced to bridge heterogeneous psychological annotations and consists of Human-Guided Linkage and Machine-Induced Consensus for constructing consistent pseudo-facet structures. Under this harmonization process, Essays and Kaggle datasets are organized into support–query pairs for prototypical learning. Two training pipelines are explored: (A) LLM-before-the-loop, where an LLM filters and refines data prior to training, and (B) LLM-in-the-loop, where the LLM dynamically assesses sample quality during training to identify ambiguous, mislabeled, or boundary cases. The resulting embeddings are attention-pooled and passed through a projection layer. The model is optimized using a meta-learning objective with contrastive loss, followed by a classification head trained with cross-entropy, enabling the formation of machine-induced pseudo-facets.}
    \label{fig:jam-framework}
    \end{figure*}

Fig.~\ref{fig:jam-framework} illustrates the proposed JAM architecture, which targets tailoring generalization across datasets annotated under different psychological theories, thus achieving a theory-agnostic model.

\subsection{Datasets}
We utilized Tan's train-validation-test split algorithm for the standard Essays and Kaggle datasets \cite{Tanfocal2025}. This stratified algorithm ensures an even distribution of personality traits across all dimensions within each split, promoting fairness in comparison and validating the effectiveness of our model. Table \ref{tab:datasets} summarizes the datasets used in this study. Note that, due to the nature of the JAM's prototypical learning, the training sets are resampled into support and query sets for each episode in training; while evaluating, the existing validation set will serve as the support set, and the existing test set will be the query set.

\begin{table*}[ht]
    \centering
    \footnotesize
    \caption{Datasets and their descriptions with the number of samples in train, validation, and test splits}
    \begin{tabularx}{\linewidth}{|p{1.9cm}|p{1cm}|p{2.75cm}|X|}
        \hline \raggedright
        \textbf{Dataset} & \textbf{Theory} &\textbf{Splitting Distribution} & \textbf{Description} \\
        \hline \raggedright
        Essays (Public) \cite{Pennebaker1999} & Big-5 & 1578, 395, 494 \newline (Train, Validation, Test) & This dataset was collected in a controlled environment where volunteers were instructed to write down whatever came to mind over a 20-minute period. It includes both self-reported ratings (via questionnaire) and projective ratings (by 18 observers).\\
        \hline \raggedright
        Kaggle (Public) \cite{mitchell2017} & MBTI & 5552, 1388, 1735 \newline (Train, Validation, Test)& This dataset comprises data crawled, labeled, and filtered from PersonalityCafe, an online community forum where users share self-reported personality test results. \\
        \hline
    \end{tabularx}
    \label{tab:datasets}
    \end{table*}

As in Eq. \ref{eq:multilabel}, we treated the dataset, $\mathcal{D}_{\text{standard}}$, as a multi-label (or multi-task) classification problem instead of a multi-class classification problem, where: $n$ is the total number of samples in the dataset, $x$ represents the sample text, and $y$ represents the labels (the number of labels depends on the dimensions of the personality theory, $d=\{d_1, d_2, \dots, d_{|d|}\}$) associated with $x$. This is because personality theory treats each dimension as independent from the others \cite{Zettler,Thielmann}. However, certain research argues that there are interdependencies or correlations between personality dimensions, suggesting that traits might not be entirely independent but could interact in complex ways \cite{Roberts,Thielmann}. Nevertheless, we adopt a multilabel classification approach to ensure that the model outputs a probability distribution over the dimensions. This approach avoids framing the task as a binary classification problem, instead allowing the intermediate layers of the neural network to automatically capture potential correlations between the labels via gradient backpropagation. In addition to the labels $y$, we include a vector of sample-specific confidence $z$, where each $z_i$ adjusts the influence of the corresponding label $y_i$ during training.

\begin{equation}
\mathcal{D}_{\text{standard}}
= \left\{(x, y, z)^n \;\left|\;
\begin{aligned}
& x= \text{sample text}\\
& y = (y_0, y_1, \dots, y_{|d|-1}), \\
& z = (z_0, z_1, \dots, z_{|d|-1}), \\
& y_i \in \{0,1\},\;
z_i \in \mathbb{R} \;\;  \forall i
\end{aligned}
\right.\right\}
\label{eq:multilabel}
\end{equation}

\subsection{Proposed Algorithms}
As illustrated in Fig.~\ref{fig:jam-framework}, the JAM framework comprises three major components designed to achieve the overall objectives. We design an \textbf{Attention-Pooled Graph Prototypical Network} to learn representations that capture underlying pseudo-facets through clustering in the embedding space. JAM incorporates a \textbf{Cross-Theory Harmonization} module, which includes \textit{Human-Guided Linkage} and \textit{Machine-Induced Consensus}, enabling the model to derive pseudo-facets from learned representations rather than relying solely on predefined human annotations, as discussed in Section~\ref{sec:pt}. We further investigate an \textbf{LLM-as-a-Judge} mechanism operating under two configurations: \textit{LLM-before-the-loop} and \textit{LLM-in-the-loop}. These configurations differ in the stage and manner in which the LLM influences training. The mechanism is incorporated to assess training sample quality and identify hard examples, including ambiguous, mislabeled, or boundary-adjacent instances, thereby guiding more focused learning.

\subsubsection{Attention-Pooled Graph Prototypical Network}

We adopt the Longformer model as the language embedding backbone due to its ability to efficiently process long textual sequences, which is particularly important for personality recognition tasks involving extensive user-generated content. Longformer employs a combination of local and global attention mechanisms, enabling scalable encoding of long documents while maintaining computational efficiency \cite{beltagy2020longformer}. Given an input text $x$, we first tokenize it and feed it into the Longformer encoder to obtain contextualized hidden representations. The encoder produces layer-wise hidden states across all transformer layers. From these, we derive a set of $L$ pooled representations, where each node corresponds to a layer-wise pooled representation of the input in $\mathbb{R}^{v}$ obtained via pooling over token embeddings within that layer. The resulting node feature matrix is defined in Eq.~\ref{eq:longformer}.

\begin{equation}
    \mathbf{H} = \text{Longformer}(x), \quad \mathbf{H} \in \mathbb{R}^{L \times v}
    \label{eq:longformer}
\end{equation}

Later, the output $\mathbf{H}$ is used as input to the graph neural network, which refines the representations by modeling interactions among embedding vectors derived from the language encoder. We construct a normalized adjacency matrix $\mathbf{\tilde{A}}$ from an adjacency matrix $\mathbf{A}$ that defines the relationships between nodes. In this work, we adopt a fully connected weighted graph, where each node is connected to all other nodes, to enable unrestricted and symmetric information exchange across all representation nodes. The number of nodes in the graph neural network corresponds to the number of layers ($L$) in the selected encoder, and the adjacency matrix remains shared across all layers. The node aggregation process is illustrated in Eq.~\ref{eq:adj}. 

\begin{equation}
\label{eq:adj}
\begin{aligned}
\mathbf{\tilde{A}} &= \mathbf{D}^{-1/2} \mathbf{A} \mathbf{D}^{-1/2}, \mathbf{D}_{ii} = \sum_j \mathbf{A}_{ij}, \\
\mathbf{A}_{ij} &= 1, \;\forall i \neq j, \mathbf{A}_{ii} = 0, \mathbf{\tilde{A}} \in \mathbb{R}^{L \times L}.
\end{aligned}
\end{equation}

This design treats all layer-wise representations as mutually interacting components without imposing predefined hierarchical or locality constraints, thereby allowing the model to learn how information should be integrated across different abstraction levels in a data-driven manner. While this choice provides a simple and uniform mechanism for cross-layer fusion, it does not explicitly encode heterogeneous or sparse inter-layer dependencies. \textbf{Investigating adaptive or learned graph structures that could more finely capture layer-specific relationships is therefore left as a promising direction for future work}.

Through matrix multiplication with the normalized adjacency matrix $\mathbf{\tilde{A}}$, information is aggregated from neighboring nodes to update node embeddings. This propagation is performed iteratively across GNN layers, where $\mathbf{H}^{(k)}$ denotes the node representations at the $k$-th layer. Eq.~\ref{eq:node_aggregation} illustrates the node update process. Here, $\sigma(\cdot)$ denotes the LeakyReLU activation function with negative slope coefficient $\alpha = 0.01$, and $\mathbf{W}^{(k)}$ is the trainable weight matrix at the $k$-th layer.

\begin{equation}
    \label{eq:node_aggregation}
    \begin{aligned}
    \mathbf{H}^{(k)} &= \sigma \left( \mathbf{\tilde{A}} \mathbf{H}^{(k-1)}
    \mathbf{W}^{(k)} \right),\\
    \sigma(x) = \max(\alpha x, &x),
    \mathbf{W}^{(k)} \in \mathbb{R}^{v \times v},
    \mathbf{H}^{(k)} \in \mathbb{R}^{L \times v}.
    \end{aligned}
\end{equation}

Lastly, we apply attention-based pooling to obtain the final graph representation, denoted as $\mathbf{h}_{\text{graph}}$, as shown in Eq.~\ref{eq:attention_pooling}. The attention mechanism computes a scalar importance score for each node based on its interaction with a trainable query vector $q$. These scores are normalized via a softmax function to obtain attention weights $\alpha_i$, which determine the contribution of each node to the final representation. This operation is fully differentiable, allowing gradients to propagate not only to the query vector $q$, but also to each dimension of the node embeddings $\mathbf{H}_i^{(k)}$. This enables the model to jointly learn which nodes are important and how the inferred embeddings should be adjusted to optimize personality trait prediction.

\begin{equation}
    \label{eq:attention_pooling}
    \begin{aligned}
    \mathbf{h}_{\text{graph}} = \sum_{i=1}^{L} \alpha_i \mathbf{H}_i^{(k)}, 
    \alpha_i = \frac{\exp\left(\mathbf{H}_i^{(k)} q\right)}{\sum_{j=1}^{L} \exp\left(\mathbf{H}_j^{(k)} q\right)}, 
    q \in \mathbb{R}^{v}
    \end{aligned}
\end{equation}

\subsubsection{Cross-Theory Harmonization (CTH)}

\begin{figure}[ht]
    \centering
    \subfigure[Regular CE]{
        \includegraphics[width=0.45\linewidth]{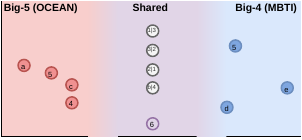}
        \label{fig:moti-ce}
    }
    \subfigure[Weighted CE]{
        \includegraphics[width=0.45\linewidth]{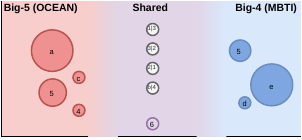}
        \label{fig:moti-weighted}
    }
    \subfigure[Prototypical Finetuning (PF)]{
        \includegraphics[width=0.45\linewidth]{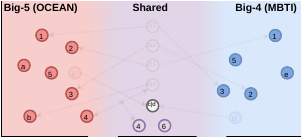}
        \label{fig:moti-pf}
    }
    \subfigure[PF + HGL]{
        \includegraphics[width=0.45\linewidth]{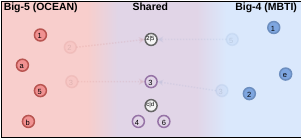}
        \label{fig:moti-HGL}
    }
    \subfigure[PF + MIC]{
        \includegraphics[width=0.45\linewidth]{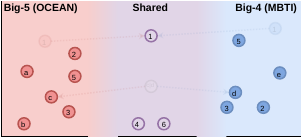}
        \label{fig:moti-MIC}
    }
    \subfigure[PF + HGL + MIC (Full CTH)]{
        \includegraphics[width=0.45\linewidth]{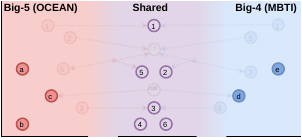}
        \label{fig:moti-CTH}
    }
    
    \caption{The conceptual schematic visualization comparing the effects of the baseline (prior) algorithm, the proposed CTH algorithm, and its ablation variants. Each particle represents a pseudo-facet that belongs to the respective cluster of the personality theory: red and blue correspond to the respective personality theories used during inference. The numbers inside each particle represent shared features that can potentially be learned and aligned. Identical numbers (e.g., 1, 2, 3, …) indicate the same feature across theories, while the numerical values themselves are only used as indices and do not carry intrinsic meaning. Alphabetic labels denote features unique to each personality theory; letters (a, b, c, …) are also meaningless sequential identifiers. Purple particles indicate shared features that have been successfully learned and aligned, while grey particles represent incorrectly aligned features. The values are presented by \textcircled{\tiny{$\cdot|\cdot$}}, where the left side corresponds to the Big-5 representation and the right side corresponds to the MBTI representation.
    }
    \label{fig:moti}
\end{figure}

In order to achieve theory-agnostic modeling, we aim to ensure that the model captures the core personality features embedded within the text. Given the presence of personality theories across different datasets, this setting provides an opportunity for the model to learn not only surface-level patterns but also to structure the representation space in a more disentangled manner. Fig. \ref{fig:moti} illustrates the overall idea behind this motivation.

Under a standard cross-entropy (CE) formulation, different personality theories are treated as independent labels. As a result, the model tends to prioritize dominant and easily separable signals during optimization, while underutilizing or completely ignoring subtler but potentially shared representational facets across theories. Although weighted CE (eg., focal loss) can rebalance gradient contributions and improve sensitivity to harder samples, these approaches primarily amplify already-learned discriminative cues rather than fundamentally restructuring the representation space. Consequently, the underlying bias toward dominant features remains largely unchanged.

To address this limitation, we adopt \textbf{prototypical fine-tuning (PF)}. The class prototypes are computed from the support set $\mathcal{D}^{(S)}$, where each prototype $s_i$ is defined as the (weighted) centroid of embeddings belonging to a given class. As shown in Eq.~\ref{eq:PF-loss}, each prototype is obtained as a weighted mean of support embeddings, where sample weights are denoted by $z$ (set to $z=1$ in our current setting). The encoder $f_\phi(x)$ maps each input $x$ into an embedding space parameterized by $\phi$, and classification is performed by comparing query embeddings against class prototypes in this space. In particular, the model minimizes the distance between a query embedding and the prototype of its ground-truth class while maximizing its distance to all other prototypes. A softmax over negative Euclidean distances is then used to define the probability of assigning a query $q$ to class $d_i$. The training objective is formulated as a weighted negative log-likelihood over the query set $\mathcal{D}^{(Q)}$, where each query is also associated with a sample-specific weight.

\begin{equation}
\begin{aligned}
    \mathcal{L}_{\text{PF}} = -w^{(Q)}& \cdot \log \frac{\exp\left(-\|f_\phi(q) - s_{i}\|^2\right)}{\sum_{j} \exp\left(-\|f_\phi(q) - s_j\|^2\right)},\; \\
    s_i &= \frac{\sum_{(x, w) \in \mathcal{D}^{(S)}_i} w \cdot f_\phi(x)}{\sum_{(x, w) \in \mathcal{D}^{(S)}_i} w}
\end{aligned}
\label{eq:PF-loss}
\end{equation}

Through sampling combinations of instances and prototypes, PF naturally alleviates data imbalance and reduces representational bias. This induces a structured clustering behavior in which distinct facets are progressively aligned with their corresponding theory-specific prototypes, while latent shared structures begin to emerge. However, at this stage, certain facets remain under-trained or ambiguously represented, which may still lead to misclassification. Overall, PF encourages the representation space to organize around theory-aware prototypes, as illustrated in Fig.~\ref{fig:moti-pf}.

We extend this with \textbf{Human-Guided Linkage (HGL)}, where external human-defined supervision is introduced to explicitly align shared facets across personality theories, encouraging the emergence of a more unified shared representation space. As in Fig. ~\ref{fig:moti-HGL}, while this strategy improves cross-theory alignment, it may also introduce noise due to imperfect or overly rigid mappings that do not always reflect true semantic correspondence. Nevertheless, these guided linkages are particularly valuable for handling edge cases: even when the provided alignments are partially inaccurate, they still gently steer the representations toward shared regions, facilitating the re-discovery and re-association of related facets.

In parallel, we explore \textbf{Machine-Induced Consensus (MIC)}. We employ a cross-entropy objective (Eq.~\ref{eq:MIC}) to guide cross-dataset adaptation via a lightweight adaptation layer that projects embeddings into a shared space. This layer is not treated as part of the core model architecture; instead, it serves as a training-time mechanism to facilitate convergence toward representations that remain discriminative across different personality theories.

\begin{equation}
    \label{eq:MIC}
\mathcal{L}_{\text{MIC}}^{(D \in \{\text{Essays}, \text{Kaggle}\})}(y, \hat{y})
= -\left[
\begin{aligned}
& y \log \hat{y} \\
& + (1 - y)\log(1 - \hat{y})
\end{aligned}
\right]
    \end{equation}

As illustrated in Fig.~\ref{fig:moti-MIC}, consensus is learned through joint optimization over paired tasks, enabling the model to infer shared structure from agreement signals rather than explicit manual supervision. This process helps refine and further ``clean'' the shared space established by prototypical fine-tuning (PF), while also revealing latent pseudo-facets that are consistently aligned across theories. However, MIC exhibits a conservative alignment tendency, prioritizing high-confidence correspondences and potentially down-weighting weaker but still informative relationships. Consequently, it is most effective when applied after Human-Guided Linkage (HGL), where it acts as a stabilizing mechanism that regularizes and consolidates the previously introduced human-guided alignments.

Lastly, we integrate PF, HGL, and MIC, termed \textbf{Cross-Theory Harmonization (CTH)}. This hybrid design leverages the complementary strengths of each component: PF provides stable prototypical anchors for organizing the representation space, HGL introduces external guidance for soft alignment of potentially shared facets across theories, and MIC further refines these relations through data-driven consensus. Together, these mechanisms progressively reshape the representation space from one dominated by isolated theory-specific signals into a coherent shared manifold, where both distinct and overlapping pseudo-facets are more faithfully encoded, and previously under-represented or unlearned features are systematically recovered and integrated.

\subsubsection{LLM-as-a-Judge (LAJ)}
While CTH aims to align representations across personality theories, residual noise may still hinder effective learning and lead to incorrect connections, as shown in Fig.~\ref{fig:moti-pf}, despite MIC efforts, which cannot fully address noise arising from mislabeled data. Moreover, while LLM-based augmentation has been shown to improve performance \cite{Tan2025picepr}, it introduces potential risks of data leakage. We therefore argue that \textbf{LLMs are better positioned as auxiliary evaluators rather than primary predictors}. In particular, encoder-based architectures remain the backbone for representation learning and classification, while LLMs are employed as reasoning-based judges for data quality assessment rather than direct personality inference. They evaluate label consistency, contradictions, and noisy or implausible samples, improving dataset integrity and reducing leakage into core predictions. This preserves the theoretical grounding of encoder-based personality modeling while leveraging LLMs' world knowledge and reasoning capabilities.

\begin{figure}[ht]
    \centering
    \fbox{\includegraphics[width=0.9\linewidth]{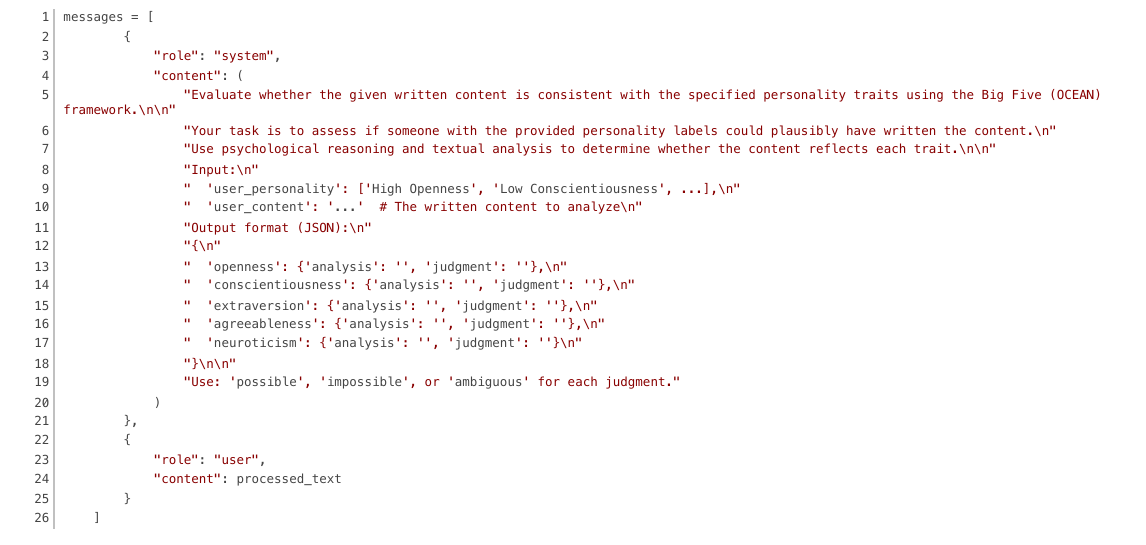}}
    \caption{The CoT System Prompt generates output in a structured JSON schema. It performs an analysis of text samples and provides recommendations on whether to `maintain', `mute,' or `lower' contributions for training.}
    \label{fig:laj}
    \end{figure}

In this work, we adopt LAJ to evaluate data correctness, specifically using (i) OpenAI's ChatGPT model (\textit{gpt}, \texttt{gpt-4o-2024-08-06}), (ii) Alibaba Qwen model (\textit{qwen}, \texttt{Qwen3.6-35B-A3B}), (iii) Meta Llama model (\textit{llama}, \texttt{Llama-3.1-8B-Instruct}), and (iv) DeepSeek model (\textit{ds}, \texttt{DeepSeek-V4-Flash}). We employ Chain-of-Thought (CoT) prompting to guide the LLMs in analyzing and determining whether a labeled sample is \textit{Possible}, \textit{Impossible}, or \textit{Ambiguous} with respect to its associated personality label, as in Fig. \ref{fig:laj}.

We further conduct QLoRA fine-tuning experiments on open-source LLMs. Training is performed exclusively on the training split using the reasoning-data synthesis method proposed by \cite{Tan2025picepr}, where ground-truth labels are leveraged to generate reasoning traces that supervise the model's decision-making process. However, the original dataset consists only of prototypical binary personality labels and lacks ambiguous cases that require reasoning over mixed or conflicting personality signals. To address this limitation, we augment the training data by constructing \textit{ambiguous} samples through the combination of two distinct personality profiles, randomly selecting sentences from each profile to form a single input. This augmentation exposes the model to more realistic borderline cases and enables us to investigate whether LLM-based judges exhibit systematic failures or introduce new prediction biases when personality evidence is ambiguous. The augmentation is applied exclusively to the training data, while the test set remains untouched, ensuring that no data leakage occurs.

As shown in Fig.~\ref{fig:jam-framework}, we study two pipelines. In the \textbf{LLM-before-the-loop (LBL)} approach (Algorithm~\ref{alg:lbl}), the LLM is employed \textit{prior to training} to assess and refine training samples based on label plausibility. In contrast, the \textbf{LLM-in-the-loop (LIL)} strategy (Algorithm~\ref{alg:lil}) integrates the LLM \textit{during training}, where it dynamically evaluates samples in real time based on their loss values.

\begin{algorithm}[ht]
\small
\caption{LLM-before-the-loop (LBL)}
\label{alg:lbl}
\begin{algorithmic}[1]
\REQUIRE $\mathcal{D}_{\text{standard}} = \{(x, y, z)\}$, where $z_i=1$ by default, system\_prompt, JSON\_output\_schema
\ENSURE Updated $\mathcal{D}_{\text{standard}}$ with modified $z$

\FORALL{$(x, y, z)$ in $\mathcal{D}_{\text{standard}}$}
    \FOR{$i = 0$ to $|d|-1$}
        \STATE Build $x$ and $y_i$ content and format as the \texttt{"user"} role, and Merge with system\_prompt
        \REPEAT
            \STATE Send to LLM using JSON\_output\_schema
            \STATE Attempt to parse LLM output
        \UNTIL{valid JSON is returned}
        \STATE Extract LLM judgment and update the corresponding $z$.
    \ENDFOR
\ENDFOR
\RETURN $\mathcal{D}_{\text{standard}}$
\end{algorithmic}
\end{algorithm}

\begin{algorithm}[ht]
\small
\caption{LLM-in-the-loop (LIL)}
\label{alg:lil}
\begin{algorithmic}[1]
\REQUIRE $\mathcal{D}_{\text{standard}} = \{(x, y, z)\}$, where $z_i = 1$ by default, model $f_\phi$, system prompt, JSON\_output\_schema, threshold $\tau$
\ENSURE Updated model parameters and possibly revised labels and weights

\FOR{each training episode}
    \STATE Sample support and query sets $(\mathcal{D}^{(S)}, \mathcal{D}^{(Q)})$ from $\mathcal{D}_{\text{standard}}$,
    \STATE Compute prototypes $\{s_j\}$ from $\mathcal{D}^{(S)}$
    \FORALL{$(x, y, w^{(Q)})$ in $\mathcal{D}^{(Q)}$}
        \STATE Compute query embedding $f_\phi(q)$ for $\mathcal{L}_{\text{JAM}^{\text{(LIL)}}}$     \IF{$\mathcal{L}_{\text{JAM}^{\text{(LIL)}}} > \tau$}
        \STATE Build $x$ and $y_i$ content and format as the \texttt{"user"} role, and Merge with system\_prompt
                        \REPEAT
                \STATE Send message to LLM with JSON\_output\_schema
                \STATE Receive and attempt to parse response
            \UNTIL{valid JSON is returned}
            \STATE Extract LLM judgment and update the corresponding $z$.
        \ENDIF
    \ENDFOR
    \STATE Aggregate and minimize total $\text{JAM}^{\text{(LIL)}}$ loss over query set
    \STATE Update model parameters $\phi$ using gradient descent
\ENDFOR
\end{algorithmic}
\end{algorithm}

Regardless of the pipeline, the outcomes, depending on the LLM's judgment, update $z$ in $\mathcal{D}_{\text{standard}}$. This updated $z$ thereby influences each example's contribution during prototype computation and training. The resulting examples are later sampled to form $\mathcal{D}_{\text{prototypical}}$ in each episode and are organized into Support ($\mathcal{D}^{(S)}$) and Query ($\mathcal{D}^{(Q)}$) sets for training. We implemented an $N$-way $K$-shot classification task. In our setting, $\mathcal{D}^{(S)} = K * N $ and $\mathcal{D}^{(Q)} = q *N$, where $N=|d|$ and $q$ depends on the GPU RAM capacity. These were designed to facilitate representational learning, enabling the model to generalize to new personality theories. Eq. \ref{eq:prototypical} illustrates the structure of the dataset for prototypical model training, while Eq. \ref{eq:sample-weight} shows the weighting in this experiment (modified later for sensitivity testing). We also used the weighted mean of the $N$ embeddings as the support set for few-shot learning.

\begin{equation}
\begin{aligned}
    &\mathcal{D}_{\text{prototype}}\\
&= \left\{
\begin{aligned}
&\mathcal{D}^{(S)} = \bigcup_{i = 1}^{N} \left\{ \left(x_{ij}^{(S)}, y_{ij}^{(S)}, z_{ij}^{(S)}\right) \;\middle|\; j = 1, \dots, K \right\}, \\
&\mathcal{D}^{(Q)} = \bigcup_{i = 1}^{N} \left\{ \left(x_{ij}^{(Q)}, y_{ij}^{(Q)}, z_{ij}^{(Q)}\right) \;\middle|\; j = 1, \dots, q \right\}
\end{aligned}
\right\}
\end{aligned}
\label{eq:prototypical}
\end{equation}

\begin{equation}
z =
\begin{cases}
1.0 & \parbox[t]{0.65\linewidth}{if $\zeta \in \{\texttt{Possible}\}$, the LLM judgment is \textbf{\textit{Possible}} (valid); maintain contributions.} \\[6pt]

0.2 & \parbox[t]{0.65\linewidth}{if $\zeta \in \{\texttt{Ambiguous}\}$, the LLM judgment is \textbf{\textit{Ambiguous}};
lower the contribution.}\\[6pt]

0 & \parbox[t]{0.65\linewidth}{if $\zeta \in\{\texttt{Impossible}\}$, the LLM judgment is \textbf{\textit{Impossible}} (invalid); mute the contribution.} 
\end{cases}
\label{eq:sample-weight}
\end{equation}

Finally, we incorporate the LLM-as-a-Judge mechanism to weight the contributions of each loss term. We then summarize the overall formulation in Eq.~\ref{eq:loss_all}, where the joint loss combines multiple components from different datasets and tasks.

\begin{equation}
\label{eq:loss_all}
\mathcal{L}_{\text{JAM}} = \phi \Big( \mathcal{L}_{\text{PF}}^{(\text{Essays})} + \mathcal{L}_{\text{PF}}^{(\text{Kaggle})} \Big) 
+ \psi \, \mathcal{L}_{\text{HGL}}^{(\text{Essays $\Leftrightarrow$ Kaggle})}
+ \rho \, \mathcal{L}_{\text{MIC}}
\end{equation}

\subsection{Experiment Design}

We train the model individually on the Essays-only dataset and the Kaggle-only dataset to establish the baseline reference for the ablation study. Next, we train on the combined dataset to evaluate whether the JAM can generalize to new personality theories. Subsequently, we study the JAM algorithm to determine whether it can improve dataset quality and model performance. The experiments use a batch size of 32, a learning rate of $1\times10^{-5}$, a maximum of 30,000 training episodes with early stopping, a random seed of 42, and 4 NVIDIA A100 GPUs. Table~\ref{tab:jap-experiment} presents the acronyms used and the corresponding experiment configurations.

\begin{table}[ht]
    \centering
    \footnotesize
    \caption{Overview of experimental configurations and notations}
\begin{tabular}{|p{1.2cm}|p{6.5cm}|}
\hline
\textbf{Notation} & \textbf{Description} \\
\hline\hline
$\text{CE}^{\dagger}$ & Regular cross-entropy serves as a standard classification baseline. \\\hline
$\text{PO}^{\dagger}$ & Off-the-shelf model as the few-shot prototypical baseline. (No training)\\\hline
$\text{PF}^{\dagger}$ & Fine-tuning the model using regular meta-learning for the few-shot prototypical baseline.($\phi=1$; $\psi=0$; $\rho=0$; $\left.\text{Eq.~\ref{eq:PF-loss}}\right|_{z=1}$) \\\hline\hline
$\text{HGL}^{\ddagger}$ & PF with HGL. ($\phi=1$; $\psi=1$; $\rho=0$; $\left.\text{Eq.~\ref{eq:PF-loss}}\right|_{z=1}$) \\\hline
$\text{MIC}^{\ddagger}$ & PF with MIC. ($\phi=1$; $\psi=0$; $\rho=1$; $\left.\text{Eq.~\ref{eq:PF-loss}}\right|_{z=1}$) \\\hline
$\text{CTH}^{\ddagger}$ & PF with (HGL + MIC). ($\phi=1$; $\psi_{e+1} \le \psi_e, \forall e \ge 1$; $\rho=1$; $\left.\text{Eq.~\ref{eq:PF-loss}}\right|_{z=1}$). \\\hline\hline
$\text{JAM}^{\text{(LBL)}}_{\textit{<model>}}$ & Proposed JAM approach with the LBL pipeline on CTH.\\\hline
$\text{JAM}^{\text{(LIL)}}_{\textit{<model>}}$& Proposed JAM approach with the LIL pipeline on CTH. \\\hline\hline
\text{[Essays]}  & Training using only the Essays Dataset. \\\hline
\text{[Kaggle]} & Training using only the Kaggle Dataset. \\\hline
\text{[Both]} & Training using both Essays Dataset and Kaggle Dataset. \\ 
\hline
\end{tabular}
    \label{tab:jap-experiment}
    \begin{minipage}{\linewidth}
\vspace{1em} 
\scriptsize
\raggedright
\textbf{Notes:}
\begin{itemize}[leftmargin=5em]
    \item[$^{\dagger}$] Baseline performance without any CTH  modules.
    \item[$^{\ddagger}$] Ablation setting experiments for CTH modules of the JAM approach.
    \item[$^{\dagger}$ $^{\ddagger}$] Experiments conducted without any weighting (no LAJ involved).
    \item[$_{\textit{<model>}}$] Indicates which LLM was used as judge in the JAM experiments.
    \item[\texttt{[·]}] Indicates the training dataset, attached after the notation.
\end{itemize}
\end{minipage}
\end{table}

\subsection{Evaluation}
To evaluate the performance, we adopted the following metrics: Regular Accuracy (RA) (Eq.~\ref{eq:jam-ra}) to determine the overall accuracy of the model, Balanced Accuracy (BA) (Eq.~\ref{eq:jam-ba}) to assess performance on imbalanced datasets, and the F1 Score (Eq.~\ref{eq:jam-f1}) to measure the model's bias. Here, $TP$, $FP$, $TN$, and $FN$ represent true positives, false positives, true negatives, and false negatives, respectively.

\begin{equation}
RA = \frac{TP + TN}{TP + TN + FP + FN}.
\label{eq:jam-ra}
\end{equation}

\begin{equation}
BA = \frac{1}{2}\left(\frac{TP}{TP + FN} + \frac{TN}{TN + FP}\right).
\label{eq:jam-ba}
\end{equation}

\begin{equation}
F1 = \frac{2 \cdot TP}{2 \cdot TP + FP + FN}.
\label{eq:jam-f1}
\end{equation}

\section{Results}

\begin{table*}[ht]
\setlength{\tabcolsep}{3.8pt}
\scriptsize
\centering
\caption{Performance comparison on the Essays dataset, including prior work, the baseline, ablation, and LAJ mechanisms.}
\begin{tabularx}{\linewidth}{|X|c|c|c|c|c|c|c|c|c|c|c|c|c|c|c|}
\hline
\multirow{2}{*}{\textbf{Experiment}}  & \multicolumn{3}{|c|}{\textbf{O - Openness}} & \multicolumn{3}{|c|}{\textbf{C - Conscientiousness}} & \multicolumn{3}{|c|}{\textbf{E - Extraversion}} & \multicolumn{3}{|c|}{\textbf{A - Agreeableness}} & \multicolumn{3}{|c|}{\textbf{N - Neuroticism}} \\
\cline{2-16}
 &      BA &      F1 &      RA &      BA &      F1 &      RA &      BA &      F1 &      RA &      BA &      F1 &      RA &      BA &      F1 &      RA \\
\hline
\hline
 Psycholinguistic MLP \cite{Mehta2020} &   &   & 0.6460 &   &   & 0.5920 &   &   & 0.6000 &   &   & 0.5880 &   &   & 0.6050\\
\hline
BERT MLP \cite{Mehta2020} & - & - & 0.6040 &   & - & 0.5730 &   & - & 0.5690 &   & - & 0.5700 &   & - & 0.5980\\
\hline
CoT with Emotion \cite{li2024eerpdleveragingemotionemotion}& - & 0.6093 & 0.6102 & - & 0.6864 & 0.6800 & - & 0.6302 & 0.6201 & - & 0.6501 & 0.6498 & - & 0.5600 & 0.5600\\
\hline
\hline
\textit{Baseline Representative [Essays]} & 0.6030 & 0.6278 & 0.6032 & 0.5346 & 0.5287 & 0.5344 & 0.6036 & 0.6117 & 0.6032 & 0.5760 & 0.6190 & 0.5749 & 0.5536 & 0.6258 & 0.5573 \\
\hline\hline
HGL [Both] & 0.5144 & 0.5482 & 0.5162 & 0.5003 & 0.4907 & 0.5000 & 0.5424 & 0.5794 & 0.5445 & 0.5575 & 0.5680 & 0.5567 & 0.5061 & 0.5396 & 0.5061 \\\hline
MIC [Both] & 0.6388 & 0.6564 & 0.6397 & 0.5426 & 0.5407 & 0.5425 & 0.5843 & 0.6019 & 0.5850 & 0.5928 & 0.5976 & 0.5911 & 0.6032 & 0.6157 & 0.6032 \\
\hline
CTH [Both] & 0.6734 & 0.6874 & 0.6741 & 0.6118 & 0.6000 & 0.6113 & 0.6278 & 0.6320 & 0.6275 & 0.6389 & 0.6341 & 0.6356 & 0.6640 & 0.6693 & 0.6640 \\
\hline\hline
\textbf{$\text{JAM}^{\text{(LBL)}}_{\textit{gpt}}$} [Both] & 0.7163 & 0.7255 & 0.7166 & 0.6422 & 0.6305 & 0.6417 & 0.6423 & 0.6424 & 0.6417 & 0.6511 & 0.6463 & 0.6478 & 0.6842 & 0.6917 & 0.6842 \\
\hline
$\text{JAM}^{\text{(LIL)}}_{\textit{gpt}}$ [Both] & 0.6752 & 0.6911 & 0.6761 & 0.6036 & 0.5934 & 0.6032 & 0.6162 & 0.6138 & 0.6154 & 0.6192 & 0.6107 & 0.6154 & 0.6781 & 0.6762 & 0.6781 \\
\hline
\end{tabularx}
\label{tab:jam-essays}
\end{table*}

\begin{table*}[ht]
\scriptsize
\centering
\caption{Performance comparison on the Kaggle dataset, including prior work, the baseline, ablation, and LAJ mechanisms.}
\begin{tabularx}{\linewidth}{|X|c|c|c|c|c|c|c|c|c|c|c|c|}
\hline
\multirow{2}{*}{\textbf{Experiment}}  & \multicolumn{3}{|c|}{\textbf{O - Openness}} & \multicolumn{3}{|c|}{\textbf{C - Conscientiousness}} & \multicolumn{3}{|c|}{\textbf{E - Extraversion}} & \multicolumn{3}{|c|}{\textbf{A - Agreeableness}} \\
\cline{2-13}
 &      BA &      F1 &      RA &      BA &      F1 &      RA &      BA &      F1 &      RA &      BA &      F1 &      RA  \\
\hline
\hline
BERT MLP \cite{Mehta2020} & - & - & 0.6840 & - & - & 0.6440 & - & - & 0.7830 & - & - & 0.7440\\
\hline
TrigNet \cite{yang-etal-2021-psycholinguistic} & - & 0.6717 & - & - & 0.6769 & - & - & 0.6954 & - & - & 0.7906 & -\\
\hline
TAE \cite{Hu_He_Wang_Zhao_Shao_Nie_2024}& - & 0.8117 & - & - & 0.7020 & - & - & 0.7090 & - & - & 0.6621 & -\\
\hline
DGCN \cite{yang2023orders} & - & 0.6719 & - & - & 0.6816 & - & - & 0.6952 & - & - & 0.8053 & -\\
\hline
\hline
\textit{Baseline Representative [Kaggle]} & 0.8163 & 0.9397 & 0.8974 & 0.7837 & 0.7392 & 0.7914 & 0.8182 & 0.7218 & 0.8720 & 0.8412 & 0.8552 & 0.8427 \\
\hline\hline
HGL [Both] & 0.5901 & 0.6717 & 0.5499 & 0.6409 & 0.5869 & 0.6398 & 0.6232 & 0.4333 & 0.6231 & 0.7583 & 0.7626 & 0.7556 \\
\hline
MIC [Both] & 0.7695 & 0.9039 & 0.8409 & 0.6902 & 0.6390 & 0.6893 & 0.6825 & 0.5010 & 0.7049 & 0.8058 & 0.8177 & 0.8058  \\
\hline
CTH [Both] & 0.8144 & 0.9260 & 0.8761 & 0.7634 & 0.7177 & 0.7660 & 0.7739 & 0.6480 & 0.8334 & 0.8408 & 0.8559 & 0.8427 \\
\hline\hline
\textbf{$\text{JAM}^{\text{(LBL)}}_{\textit{gpt}}$} [Both]& 0.8133 & 0.9287 & 0.8801 & 0.7721 & 0.7280 & 0.7735 & 0.7872 & 0.6659 & 0.8403 & 0.8498 & 0.8648 & 0.8519 \\
\hline
$\text{JAM}^{\text{(LIL)}}_{\textit{gpt}}$ [Both]& 0.7941 & 0.9194 & 0.8651 & 0.7538 & 0.7072 & 0.7556 & 0.7356 & 0.5818 & 0.7879 & 0.8171 & 0.8372 & 0.8202 \\
\hline
\end{tabularx}
\label{tab:jam-kaggle}
\end{table*}

\subsection{Baseline Acquisition}
\begin{figure}[ht]
    \centering
    \subfigure[Essays Dataset]{
        \includegraphics[width=1\linewidth]{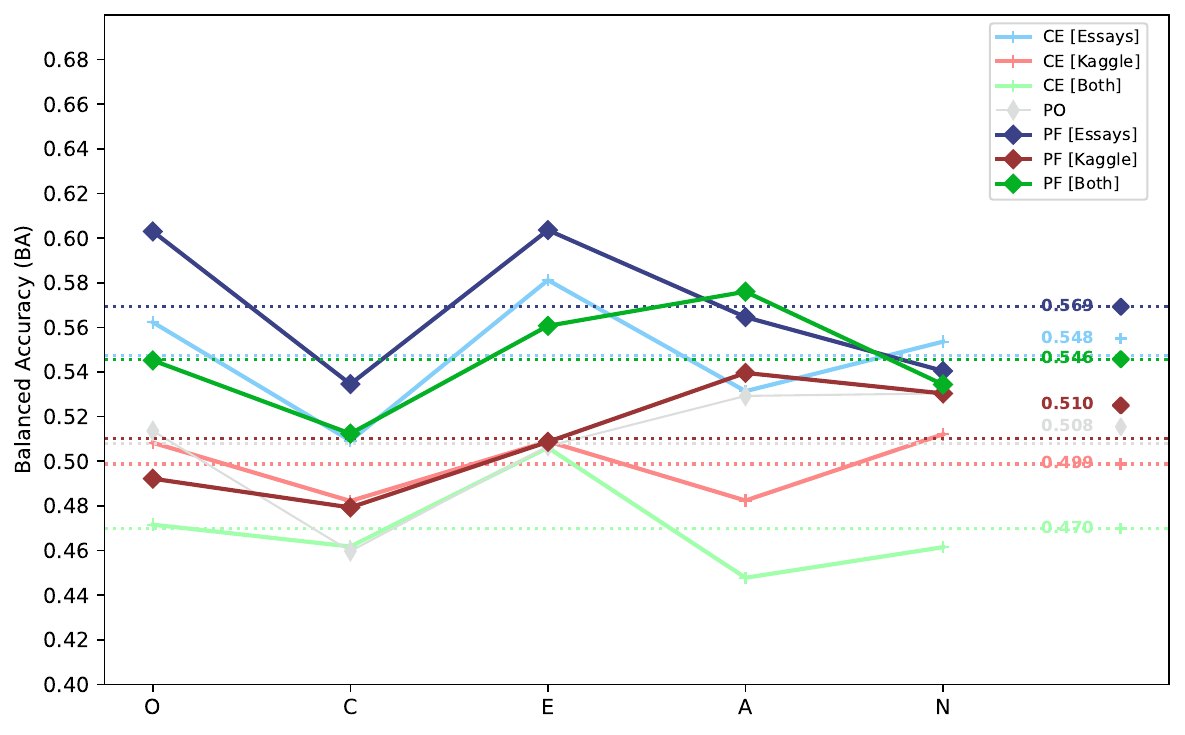}
        \label{fig:essays_jam_base}
    }
    \subfigure[Kaggle Dataset]{
        \includegraphics[width=1\linewidth]{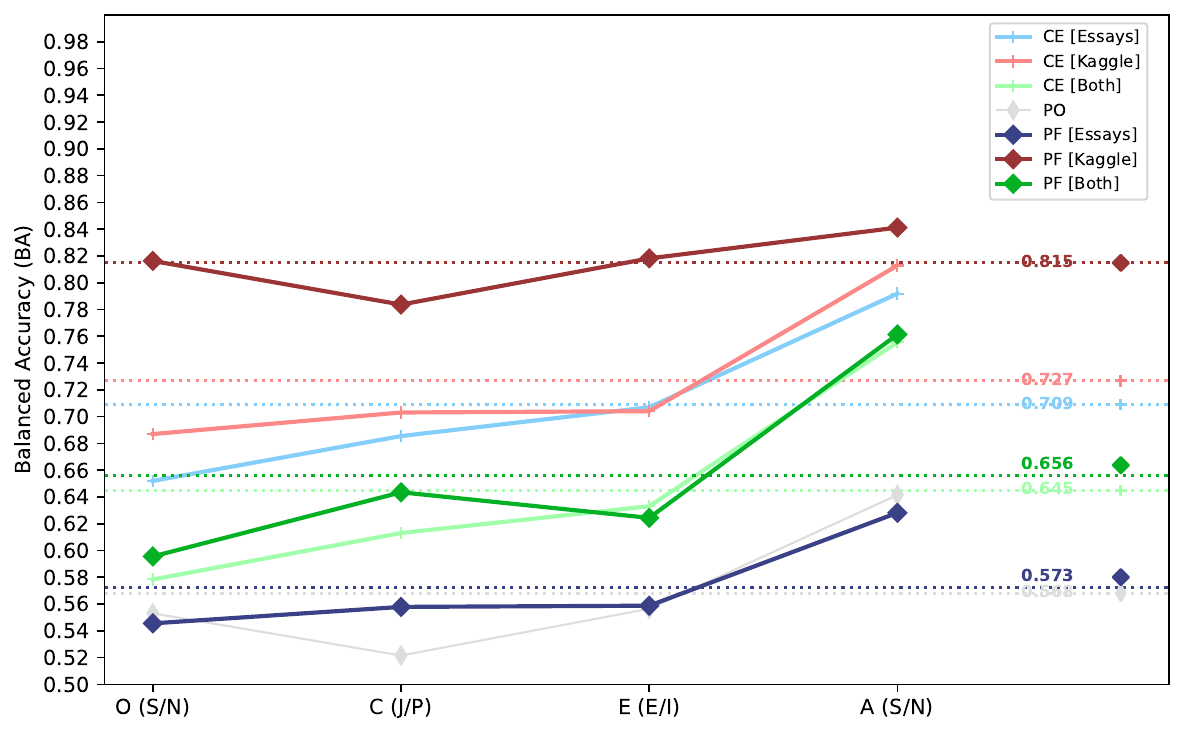}
        \label{fig:kaggle_jam_base}
    }
    \caption{Visualization of balanced accuracy for regular classification using cross-entropy (CE), prototypical few-shot learning using off-the-shelf ready model (PO), and fine-tuned prototypical few-shot learning (PF) on the Essays and Kaggle datasets. Each line in the legend corresponds to an individual experiment, with the dataset used for training indicated in the respective brackets, while the dotted lines with numbers represent the average value across dimensions of the respective experiment.
    }
    \label{fig:jam-baseline}
\end{figure}

First, we establish a baseline for comparison by training the model separately on the Essays dataset, the Kaggle dataset, or both. Fig.~\ref{fig:jam-baseline} illustrates the balanced accuracy of each approach under different dataset settings: Cross-Entropy (CE), Off-the-shelf Prototypical (PO), and Fine-tuned Prototypical (PF).

In the CE experiment, there is no doubt that when the model is trained and evaluated on the same personality theory dataset, it achieves relatively good performance. However, when the model is trained on one dataset and evaluated on another, there is a significant drop in performance, indicating that the model struggles to generalize across different personality theories. We also observe that a combination of datasets does not improve performance; rather, it degrades it. This is likely because the model is confused by the conflicting signals from the two different personality theories.

On the other hand, in the PO experiment, we observe that the Essays dataset shows very poor performance, while the Kaggle dataset is still able to capture relationships. This can be attributed to the nature of how the datasets are collected. The Essays dataset comes from a constrained environment, whereas the Kaggle dataset is sourced from social media, which is more informal and diverse. This provides confidence that few-shot learning can be effective in personality recognition tasks. One interesting observation is that it naturally solves the class imbalance problem, as the algorithm focuses on learning the class prototypes rather than being biased towards the majority class.

When the model is fine-tuned using PF, we see a significant improvement in performance within the same dataset (training and evaluation). This is especially evident for the Kaggle dataset, giving confidence that the model can learn the underlying personality patterns in the data. An interesting observation is that a model trained on the Kaggle dataset can generalize to the Essays dataset, but not vice versa, further demonstrating the quality of the Kaggle dataset. Additionally, the results show that the model trained on the Kaggle dataset achieves similar results to the CE approach trained on the Essays dataset, further justifying that a prototypical network has potential in capturing personality features from text.

\subsection{Performance}

\subsubsection{Cross-Theory Harmonization (CTH) Performance} 

Table \ref{tab:jam-essays} and Table \ref{tab:jam-kaggle} tabulate the performance of each CTH module under different dataset settings with its ablation study. To provide a reference baseline, we include the row corresponding to the highest performance (BA) from the aforementioned experiments (as shown in Fig. \ref{fig:jam-baseline}) as representative baseline. We also include prior work; however, it is not fully comparable since they focus on training on a single dataset, whereas we train 1 model for 2 personalities theories. Nevertheless, our performance remains competitive and significant in most cases, highlighting the advantage of our approach for generalization.

By observing the results on the Essays dataset, we find that the model improves on average by more than 9\% in balanced accuracy across all traits compared to the best-performing baseline. We further analyze the contribution of each component. Across both datasets, when only HGL is included, performance actually degrades regardless of the dataset. This supports the aforementioned hypothesis that human knowledge is limited; in the Essays dataset, due to its constrained data collection environment, such limitations may hinder the model and expose it to suboptimal supervision, obscuring the ultimate learning objective through self-consensus. Considering MIC alone, it only outperforms the baseline on the Essays dataset but not on Kaggle, further strengthening the observation that the combination of Essays data introduces additional noise. Nevertheless, the combination of both components (CTH) is able to restore performance to a level comparable with the baseline on Kaggle and further improve results on the Essays dataset, suggesting that integrating HGL and MIC allows the model to better merge complementary knowledge sources, achieving theory-agnostic modeling.

\subsubsection{LLM-as-a-Judge (LAJ) Performance}

Next, we study the impact of the LAJ mechanism. Generally, the $\text{JAM}^{\text{(LBL)}}_{\textit{gpt}}$ [Both] approach consistently outperforms the $\text{JAM}^{\text{(LIL)}}_{\textit{gpt}}$ [Both] approach across all personality traits in both datasets. This suggests that pre-evaluating and refining the dataset before training is more effective than dynamically assessing samples during training. The $\text{JAM}^{\text{(LBL)}}_{\textit{gpt}}$ [Both] model converges within approximately 3000 episodes, whereas $\text{JAM}^{\text{(LIL)}}$ [Both] requires around 7000 episodes to converge. This difference arises because the $\text{JAM}^{\text{(LIL)}}_{\textit{gpt}}$ [Both] approach only gradually obtains a cleaner dataset over multiple iterative episodes, and not all samples are evaluated by the LLM due to the thresholding mechanism applied during training. As a result, an initially noisy dataset may mislead the model toward suboptimal local minima, which helps explain why the $\text{JAM}^{\text{(LIL)}}_{\textit{gpt}}$ [Both] approach is less effective than $\text{JAM}^{\text{(LBL)}}_{\textit{gpt}}$ [Both]. This trend is also reflected in the Essays dataset, where the performance of $\text{JAM}^{\text{(LBL)}}_{\textit{gpt}}$ [Both] is comparable to $\text{CTH}$ [Both].

\begin{figure}[ht]
    \centering
    \subfigure[Essays Dataset]{
        \includegraphics[width=0.45\linewidth]{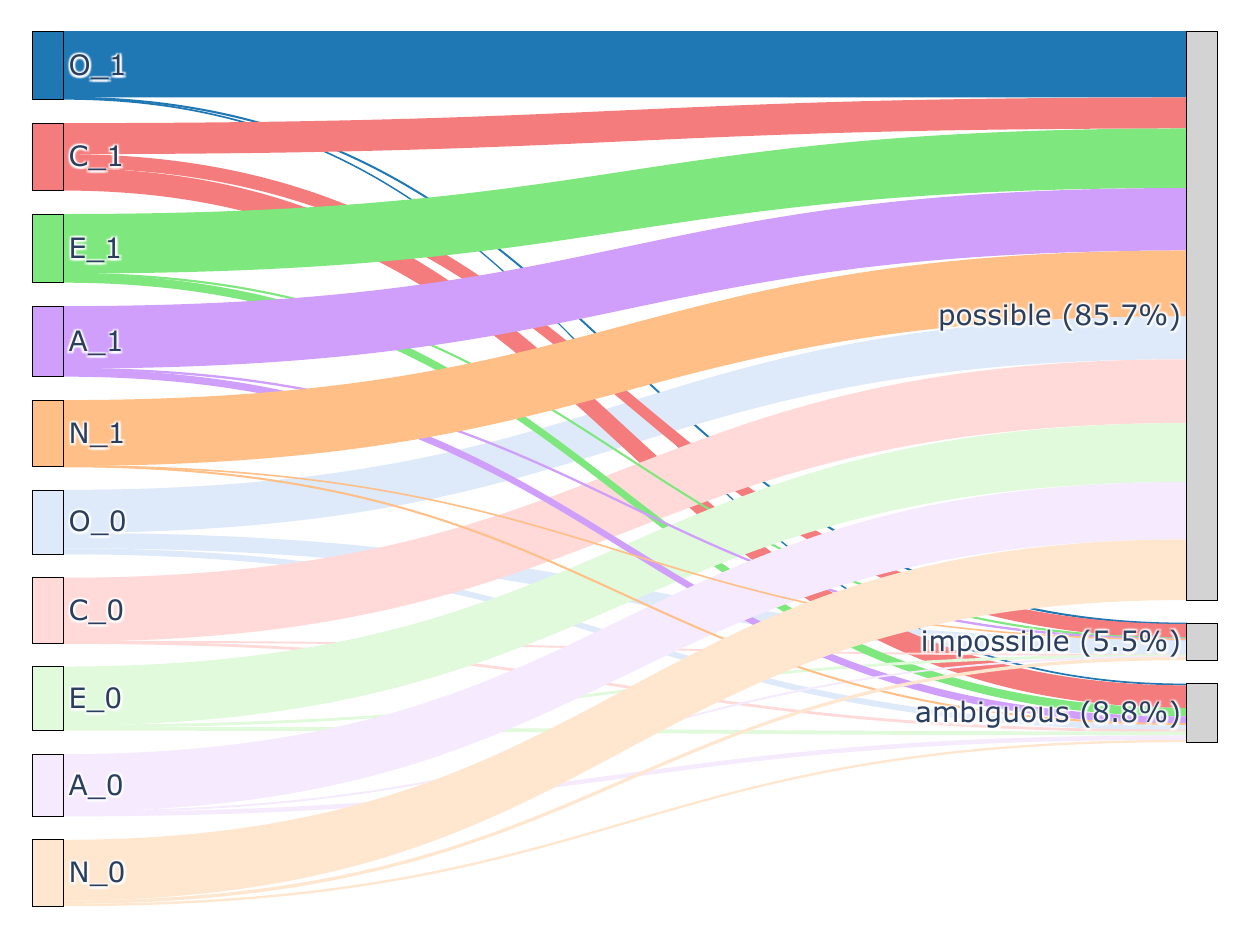}
        \label{fig:essays_dis}
    }
    \subfigure[Kaggle Dataset]{
        \includegraphics[width=0.45\linewidth]{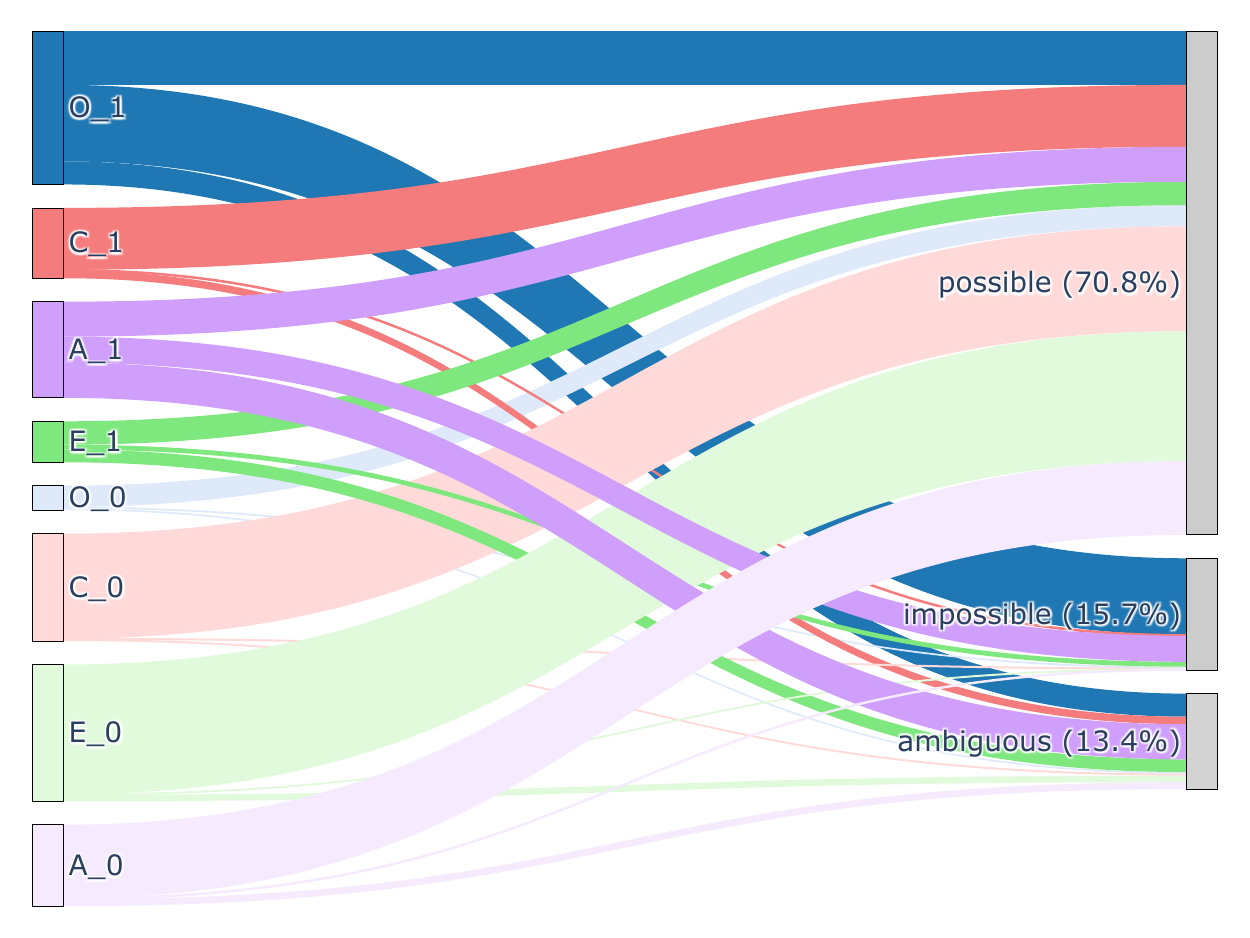}
        \label{fig:kaggle_dis}
    }
    
    \caption{
        Distribution of $\text{LLM}_{\text{gpt}}$ judgments on personality trait implications across datasets. The figure illustrates the proportion of LLM-judged outcomes—\textit{Possible}, \textit{Impossible}, and \textit{Ambiguous}—based on the binary labels for each of the personality traits. Labels in the form \text{d\_y} represent trait \text{d} with label \text{y}, where \text{y = 1} denotes a positive class (presence of the trait) and \text{y = 0} denotes a negative class. The divergence in distributions highlights the influence of annotation conditions and data origin on interpretability judgments made by language models.
    }
    \label{fig:dataset-judgement}
\end{figure}

Fig.~\ref{fig:dataset-judgement} illustrates the distribution of \textit{Possible} and \textit{Impossible} judgments across personality traits for both datasets. The Essays dataset contains a higher proportion of \textit{Possible} judgments (85.7\%) compared to the Kaggle dataset (70.8\%). Given that only a limited proportion of the dataset is estimated to be noisy (approximately $>8\% $), the observed 2\% improvement is within a reasonable range. This indicates that the filtering process effectively reduces the influence of noisy samples, while also suggesting that the achievable performance gain is naturally bounded by the proportion of removable noise in the dataset.

\begin{figure}[ht]
    \centering
    \subfigure[Essays Dataset]{
        \includegraphics[width=\linewidth]{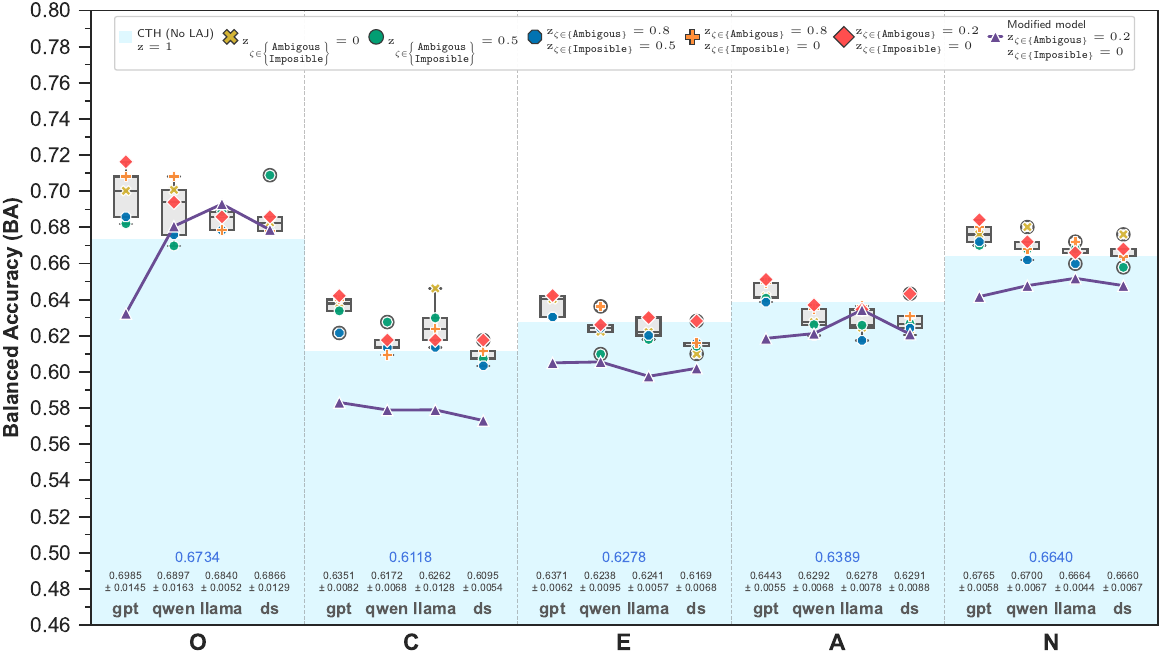}
        \label{fig:essays-llm}
    }
    \subfigure[Kaggle Dataset]{
        \includegraphics[width=\linewidth]{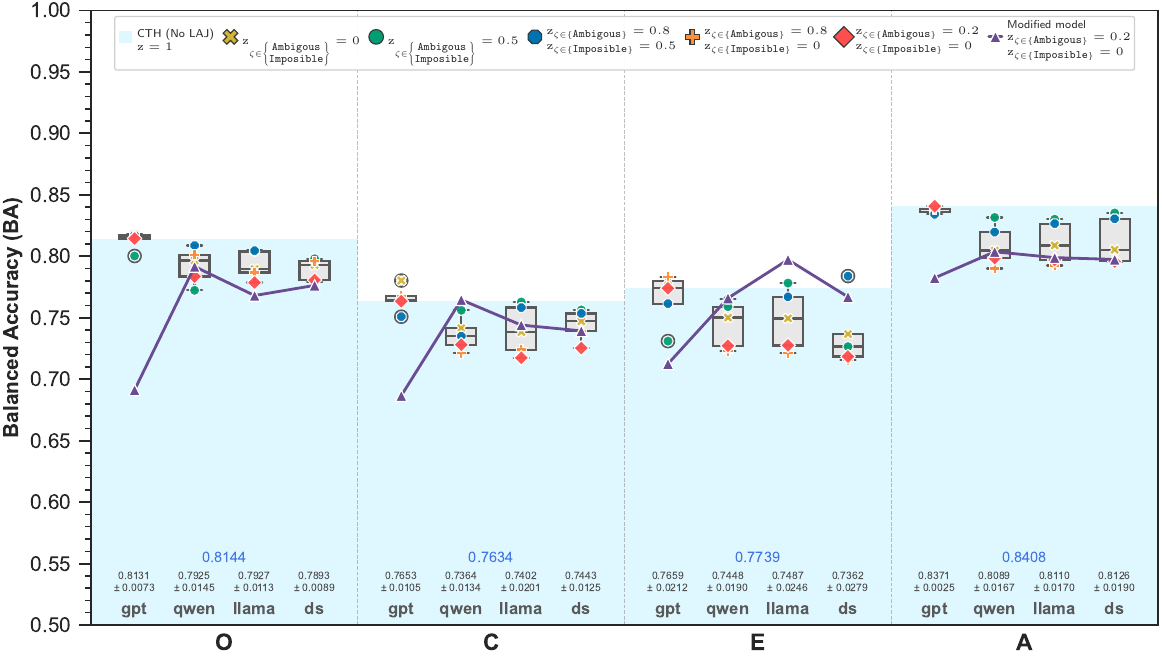}
        \label{fig:kaggle-llm}
    }
    \caption{Visualization of different large language models (\textit{gpt}, \textit{qwen}, \textit{llama}, \textit{ds}) and their performance under respective hyperparameter settings on $\zeta$ to $z$ values in $\text{JAM}^{\text{(LBL)}}_{\textit{gpt}}$ [Both], including a reduced contribution of the \texttt{ambiguous} and \texttt{impossible} flags by down-weighting them at certain levels. We further incorporate fine-tuning for open-source models using QLoRA, while for \textit{gpt} we conduct an experiment in which 80\% of the minor (relatively underrepresented) labels/classes are removed to address class imbalance, under the assumption that the LLM significantly underperforms on the task, in order to study the potential effects of such imbalance induction.
    }
    \label{fig:llm-compare}
\end{figure}

Fig.~\ref{fig:llm-compare} shows the performance of different LLMs under varying hyperparameter settings. Overall, \textit{gpt} performs the best. On the Kaggle dataset, LLMs generally do not improve over the baseline, which is consistent with the CTH stage (without LAJ). In contrast, the Essays dataset shows a different trend: with the involvement of LAJ, performance generally improves over CTH alone. This supports the generalisability claims for the low-resource theory (Essays dataset) compared with the Kaggle dataset, showing that in most cases, model selection does not degrade performance.

We observed that setting $z_{\zeta \in \{\texttt{Impossible}\}} = 0$ (muting the \textit{Impossible} samples) often leads to better performance in the Essays dataset, supporting the generalisation claim. However, this effect is less consistent in the Kaggle dataset and depends more on the choice of LLM. From the results of the Kaggle dataset, $z_{\zeta \in \{\texttt{Impossible}\}} = 0$ generally has relatively low performance compared to other hyperparameters, suggesting that the model is not only incapable of filtering data but also introduces more noise by removing some important contributions. In addition, we conduct an experiment in which 80\% of the minor (relatively underrepresented) labels/classes are removed (in \textit{gpt}) using the setting $z_{\zeta \in \{\texttt{Ambiguous}\}} = 0.2$ and $z_{\zeta \in \{\texttt{Impossible}\}} = 0$, to simulate a worse model scenario. The results show a significant drop across all dimensions, despite this being the best observed hyperparameter setting for \textit{gpt}. This again supports the claim that a worse LLM can negatively affect training, and shows that $z_{\zeta \in \{\texttt{Impossible}\}}$ is sensitive.

Across different models, $z_{\zeta \in \{\texttt{Ambiguous}\}}$ does not significantly affect performance (evaluated under settings $z_{\zeta \in \{\texttt{Ambiguous}\}} = 0.2$ and $z_{\zeta \in \{\texttt{Impossible}\}} = 0.8$, as well as $z_{\zeta \in \{\texttt{Impossible}\}} = 0$). This suggests that it primarily acts as a tunable non-sensitive hyperparameter rather than a structural factor. On the other hand, in terms of their fine-tuned model, the results show that although there is improvement in certain dimensions for some LLMs on the Kaggle dataset, the gains are not statistically significant. In some cases, performance even drops on the Essays dataset, likely due to imbalance, as Kaggle is more dominant and relatively easier to learn. These findings suggest that well-trained LLMs are already sufficiently capable, and additional fine-tuning is unnecessary and induces higher costs for the judging task.

\subsection{Statistical Findings}

\begin{figure}[ht]
    \centering
    \subfigure[\tiny\shortstack{Essays Dataset\\(PF$\rightarrow$MIC)[Both]}]{
        \includegraphics[width=0.29\linewidth]{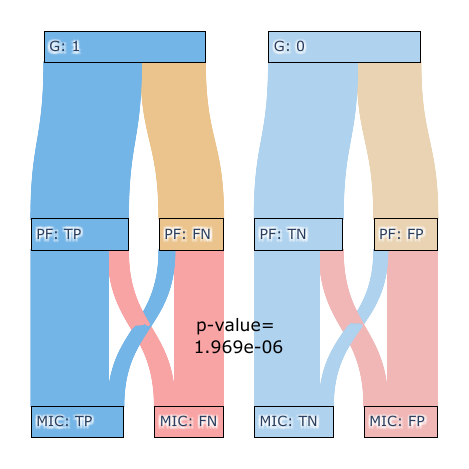}
        \label{fig:essays-pf-mic}
    }
    \subfigure[\tiny\shortstack{Essays Dataset\\(MIC$\rightarrow$CTH)[Both]}]{
        \includegraphics[width=0.29\linewidth]{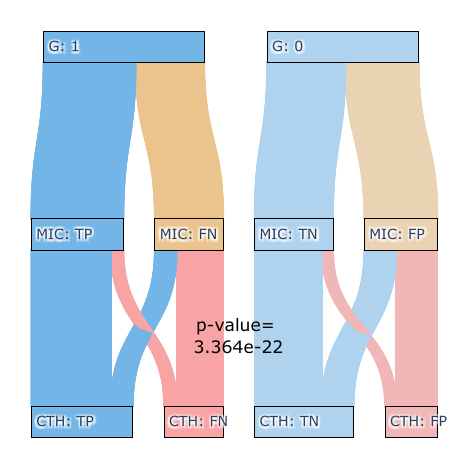}
        \label{fig:essays-mic-cth}
    }
    \subfigure[\tiny\shortstack{Essays Dataset\\(CTH$\rightarrow\text{JAM}^{\text{(LBL)}}_{\textit{gpt}}$)[Both]}]{
        \includegraphics[width=0.29\linewidth]{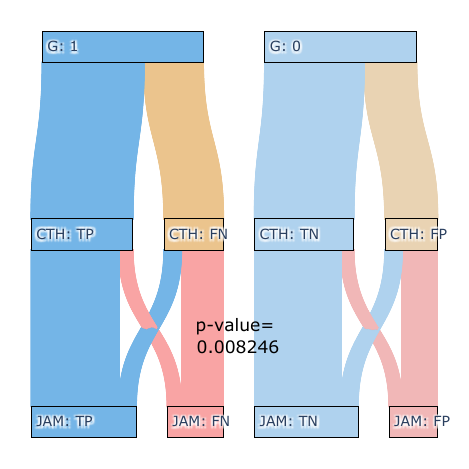}
        \label{fig:essays-cth-jam}
    }
    \subfigure[\tiny\shortstack{Kaggle Dataset\\(PF$\rightarrow$MIC)[Both]}]{
        \includegraphics[width=0.29\linewidth]{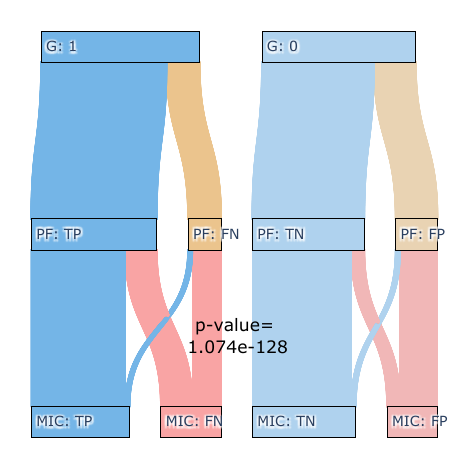}
        \label{fig:kaggle-pf-mic}
    }
    \subfigure[\tiny\shortstack{Kaggle Dataset\\(MIC$\rightarrow$CTH)[Both]}]{
        \includegraphics[width=0.29\linewidth]{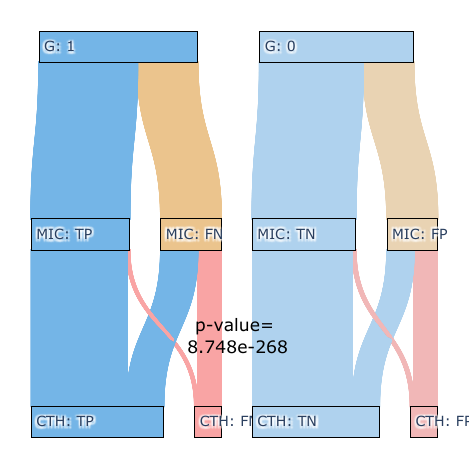}
        \label{fig:kaggle-mic-cth}
    }
    \subfigure[\tiny\shortstack{Kaggle Dataset\\(CTH$\rightarrow\text{JAM}^{\text{(LBL)}}_{\textit{gpt}}$)[Both]}]{
        \includegraphics[width=0.29\linewidth]{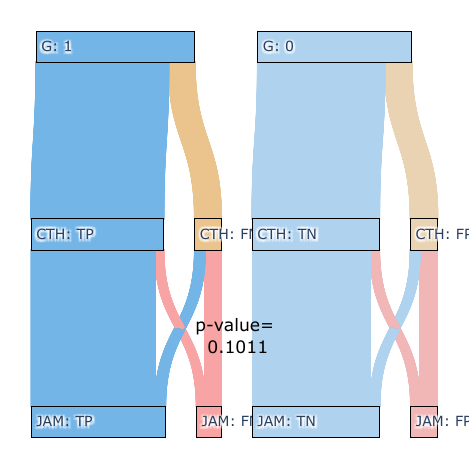}
        \label{fig:kaggle-cth-jam}
    }
    
    \caption{The Sankey diagram illustrates the transitions between 4 approaches: from the PF [Both] to the MIC-involved method (first MIC [Both], and then to CTH [Both] method, and to the proposed full approach $\text{JAM}^{\text{(LBL)}}_{\textit{gpt}}$ [Both]). Note that the HGL-only is excluded from this test since it functions as early generalization guidance. It encapsulates noise and performs worse when functioning alone. Each method’s outcomes are compared against the original ground truth (G). Statistical significance is determined using the McNemar $p$-value test. Due to the fact that personality encompasses multiple dimensions, we flatten and concatenate these dimensions to facilitate clearer visualization.}
    \label{fig:jam-stats}
\end{figure}

We conducted the McNemar test to support our findings, as illustrated in Fig. \ref{fig:jam-stats}. The results indicate its potential utility in constrained environments, while the CTH method addresses the challenges of merging datasets, suggesting a pathway toward a theory-agnostic model. Furthermore, this demonstrates that our proposed approaches are able to reduce noise that appeared in the native collection of the dataset, where it statistically further improves through LAJ (an average of 2\% improvement in the Essays dataset).

\begin{figure}[ht]
    \centering
    \subfigure[\shortstack{Essays Dataset \\PO [Essays]}]{
        \includegraphics[width=0.45\linewidth]{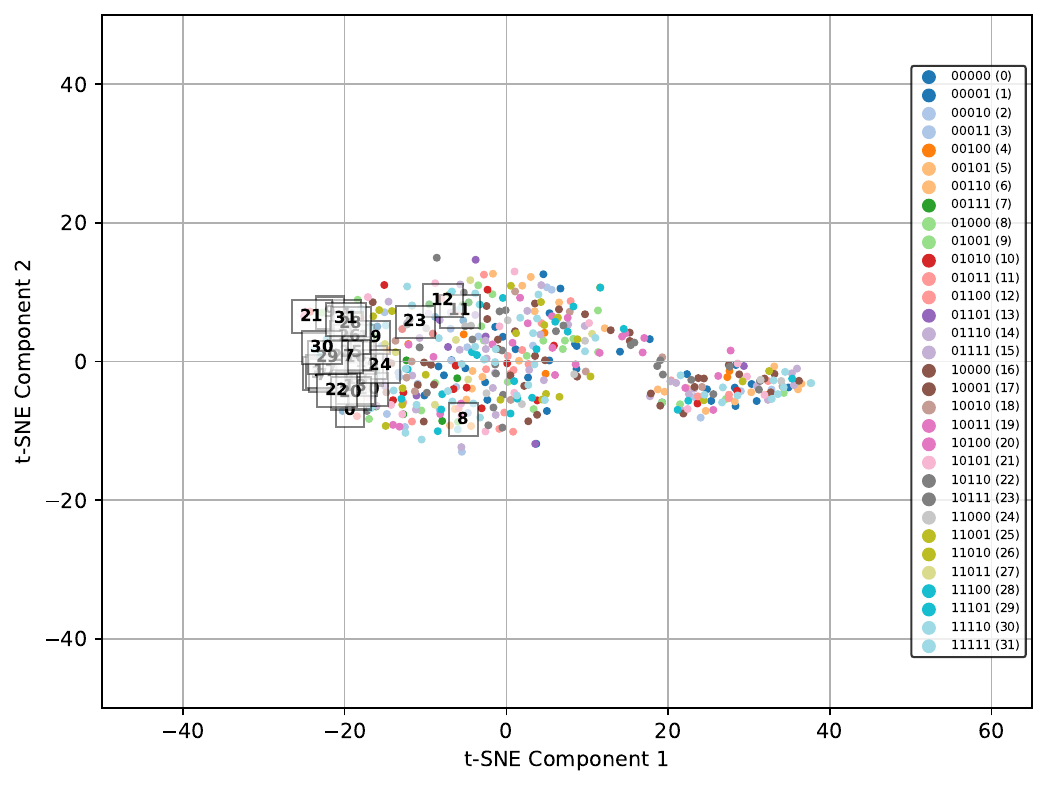}
        \label{fig:essays_jam_s_d}
    }
    \subfigure[\shortstack{Essays Dataset \\ $\text{JAM}^{\text{(LBL)}}_{\textit{gpt}}$[Both]}]{
        \includegraphics[width=0.45\linewidth]{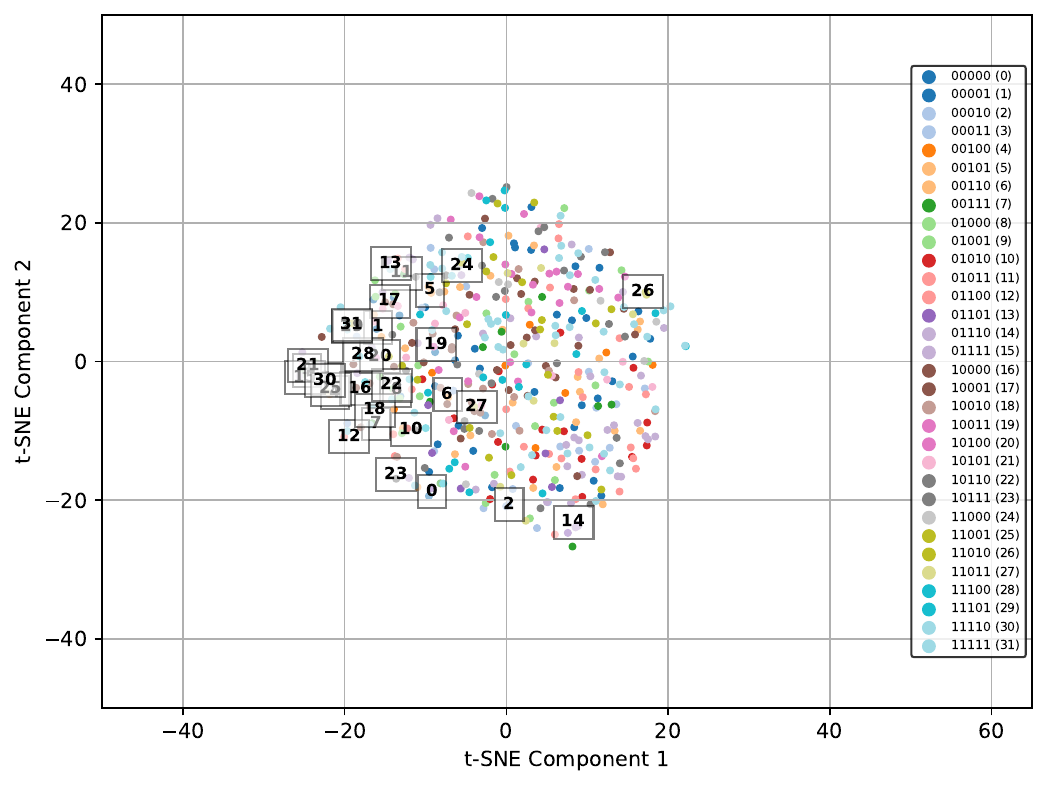}
        \label{fig:essays_jam_s_p}
    }
    \subfigure[\shortstack{Kaggle Dataset \\PO [Kaggle]}]{
        \includegraphics[width=0.45\linewidth]{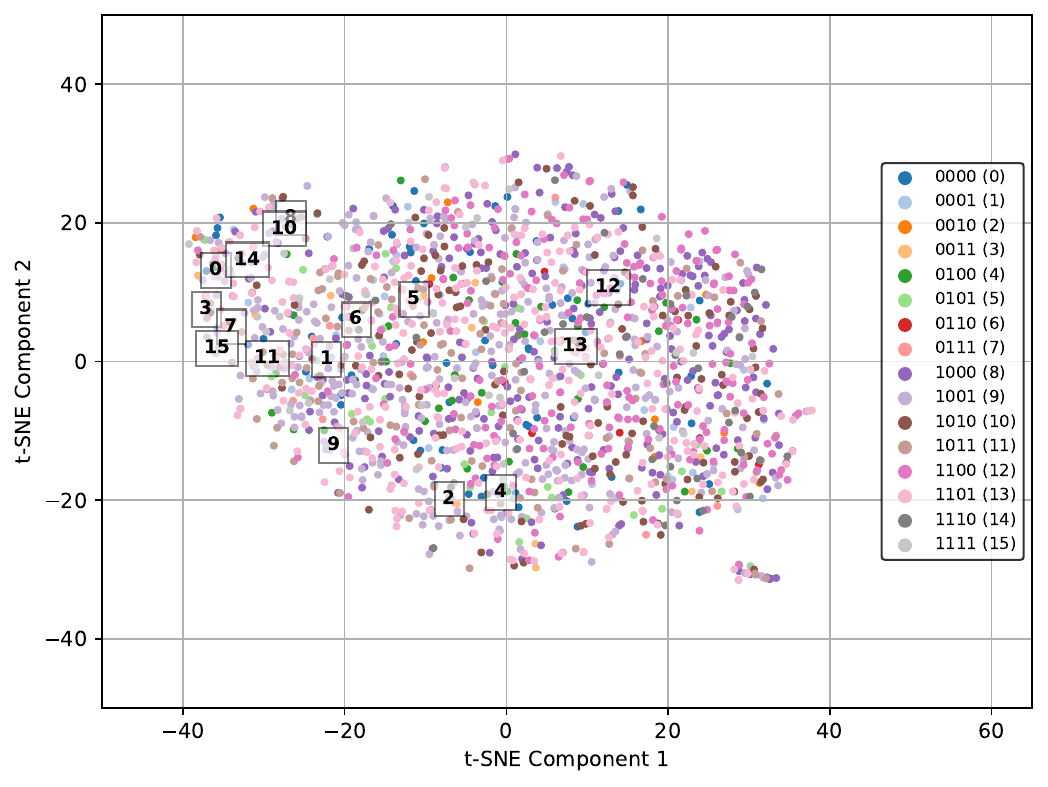}
        \label{fig:kaggle_jam_s_d}
    }
    \subfigure[\shortstack{Kaggle Dataset \\$\text{JAM}^{\text{(LBL)}}_{\textit{gpt}}$[Both]}]{
        \includegraphics[width=0.45\linewidth]{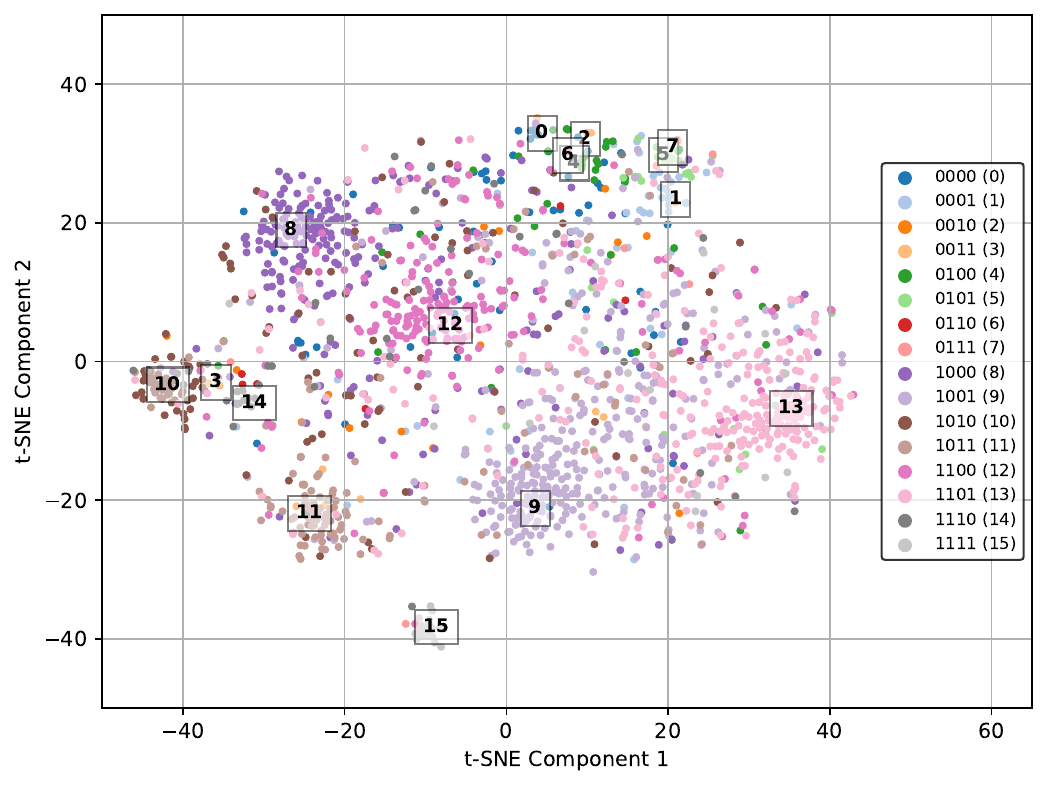}
        \label{fig:kaggle_jam_s_p}
    }
    
    \caption{The t-SNE visualization of personality embeddings on both Essays and Kaggle datasets. Subplots \ref{fig:essays_jam_s_d} and \ref{fig:kaggle_jam_s_d}  show embeddings using the off-the-shelf model (PO), while subplots \ref{fig:essays_jam_s_p} and \ref{fig:kaggle_jam_s_p} show embeddings using the proposed $\text{JAM}^{\text{(LBL)}}_{\textit{gpt}}$ [Both] approach. Each color represents a distinct personality trait combination, with annotated labels indicating cluster centers. Compared to the PO approach, $\text{JAM}^{\text{(LBL)}}_{\textit{gpt}}$ [Both] produces embeddings that are more structured and better separated, reflecting improved modeling of personality cues obtained from text.}
    \label{fig:jam-tsne}
\end{figure}

To visualize the effectiveness of the proposed algorithm, we visualized the personality embeddings using t-SNE in Fig. \ref{fig:jam-tsne}. Compared to the off-the-shelf PO approach, the embeddings produced by $\text{JAM}^{\text{(LBL)}}_{\textit{gpt}}$ [Both] are noticeably more structured and exhibit clearer separation between clusters corresponding to distinct personality trait combinations. This improved separation indicates that the proposed method captures and preserves subtle personality cues from textual data more effectively. In particular, the cluster centers in the $\text{JAM}^{\text{(LBL)}}_{\textit{gpt}}$ [Both] visualizations are more distinct and less overlapping, suggesting that the model can better differentiate between similar personality profiles. This pattern is consistent across both datasets, with the effect being particularly pronounced in the Kaggle dataset, where clusters are clearer due to a larger number of samples and relatively higher classification accuracy, further highlighting the robustness and generalizability of $\text{JAM}^{\text{(LBL)}}_{\textit{gpt}}$ [Both].

\subsection{Computational Analysis}

\begin{figure}[ht]
    \centering
    \includegraphics[width=\linewidth]{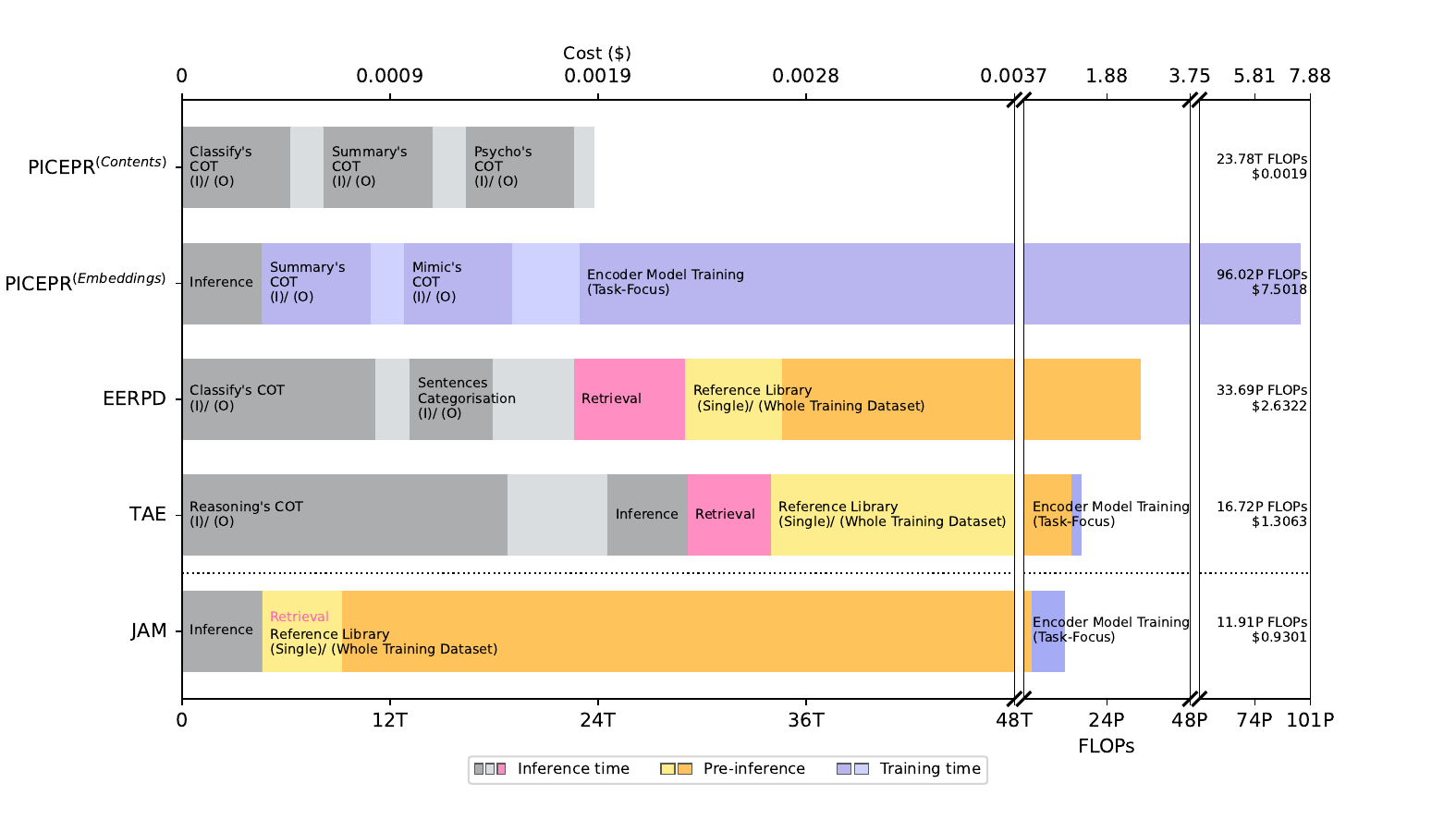}
    \caption{Visualization of Cost and FLOPs Comparison. In this analysis, we assume that input and output tokens incur identical costs, adopting the estimated FLOPs per token and the corresponding pricing model used in \cite{Tan2025picepr} to enable direct comparison. We assume a constant inference cost of 3.2 GFLOPs per token and a pricing rate of \$0.25 per 10M tokens. The visualization is based on three components: (i) single inference (grey, with pink representing search time, e.g., retrieval-augmented generation lookup or similarity measurement, quantified according to the respective approach used to achieve the reported performance), (ii) training time (purple, covering all experiments), and (iii) pre-inference (yellow and orange). Pre-inference (e.g., retrieval-augmented generation indexing or prototype preparation) is difficult to standardize because greater involvement can improve performance for certain methods. To address this, we divide the pre-inference stage into yellow and orange segments to represent an estimated ratio of involved classes or prototypes, where this is quantified according to the respective approach used to achieve the reported performance. Note that this visualization excludes the training cost of the pretrained decoder-only model, as it is common across all approaches and assumed to be comparable.}
    \label{fig:cost}
    \end{figure}

Fig.~\ref{fig:cost} compares the proposed JAM with previous approaches in terms of Floating Point Operations (FLOPs) and the corresponding computational cost. This comparison applies to both \textbf{LLM-before-the-loop (LBL)} and \textbf{LLM-in-the-loop (LIL)} settings. Technically, the former achieves lower computational cost because the filtering mechanism depends only on the loss values; therefore, not all samples in the dataset require full inference.

Overall, the proposed method achieves the lowest inference time compared to the $\text{PICEPR}^{(\mathrm{Embeddings})}$ method \cite{Tan2025picepr}. Although a slight additional overhead is introduced due to prototype retrieval, this overhead is negligible because it only involves retrieving and averaging 4--5 embedding vectors. Despite this, the method still achieves approximately $8\times$ lower training FLOPs. Furthermore, the proposed approach demonstrates the feasibility of incorporating LLMs to improve performance, while simultaneously reducing the risk of data leakage, which is difficult to achieve with purely decoder-only model-driven approaches such as $\text{PICEPR}^{(\mathrm{Contents})}$ \cite{Tan2025picepr}. Overall, JAM maintains very low inference latency, which is particularly important for test-time deployment and real-world inference scenarios, while also reducing inference FLOPs, thereby lowering energy consumption and carbon footprint and improving the environmental sustainability especially on large-scale deployment.

\subsection{Limitations and Future Works}

Although JAM demonstrates consistent improvements across heterogeneous personality datasets, it is currently validated on only two personality theories (MBTI and Big-5), and further evaluation on additional theories and culturally diverse datasets is needed to validate the robustness of the proposed Cross-Theory Harmonization (CTH). In addition, as personality recognition relies on sensitive personal data, future work should prioritize privacy, informed consent, and fairness. One promising direction is to integrate JAM with a federated learning framework, enabling collaboration across institutions or countries while keeping data local and aggregating only model updates, thereby improving generalizability without compromising data privacy.

\section{Conclusion}

In this work, we introduced Large-Language-Models-as-a-Judge in Theory-Agnostic Adaptive Metric-Alignment for Prototypical Networks in Personality Recognition (JAM), a novel \textbf{prototypical framework} designed to enhance personality prediction and generalizability by leveraging \textbf{Cross-Theory Harmonization} (\textit{Human-Guided Linkage} and \textit{Machine-Induced Consensus}) and \textbf{LLM-as-a-Judge} mechanisms, enabling the learning of latent pseudo-facets that capture shared behavioral structure across heterogeneous personality theories under an embedding space.

Through experiments on the Essays and Kaggle datasets, we demonstrated that JAM consistently outperforms existing baselines and prior work across multiple personality traits, with no class imbalance issues. Specifically, JAM achieves an average BA improvement of 12\% on the Essays dataset and 14\% on the Kaggle dataset compared to the regular prototypical few-shot learning approach. When combining datasets, it becomes possible to leverage the dataset for generalization in low-resource situations, where the performance of the Kaggle dataset is not affected. Moreover, incorporating the LLM-as-a-Judge mechanism further improves performance on the Essays dataset by an additional 2.4\%, by reweighting the contribution of each data row during model training. These results indicate that JAM is particularly effective in constrained environments while maintaining robust performance across diverse datasets. 

\section{Acknowledgments}
This research was funded by the Universiti Tunku Abdul Rahman Research Fund (IPSR/RMC/UTARRF/2021-C1/K03). The authors also appreciate the support of Grid5000, France, for providing the computational resources used in this study. The authors thank the Advanced Artificial Intelligence Research Center at the Kanagawa Institute of Technology, Japan, for supporting OpenAI's model inferences. The first author, Jing Jie Tan, also appreciates National Yang Ming Chiao Tung University for providing the Research Scholarship to establish a professional connection for guidance related to psychology. Additionally, he appreciate the Embassy of France to Malaysia for the Doctoral Research Mobility Grant and the Japan Student Services Organization (JASSO) Scholarships, which facilitated the research collaboration between Universiti Tunku Abdul Rahman, Malaysia, Université Sorbonne Paris Nord, France, and Kanagawa Institute of Technology, Japan.

\bibliographystyle{IEEEtran}
\bibliography{references}

\section*{Author Biographies}

\begin{IEEEbiography}[{\includegraphics[width=1in,height=1.25in,clip,keepaspectratio]{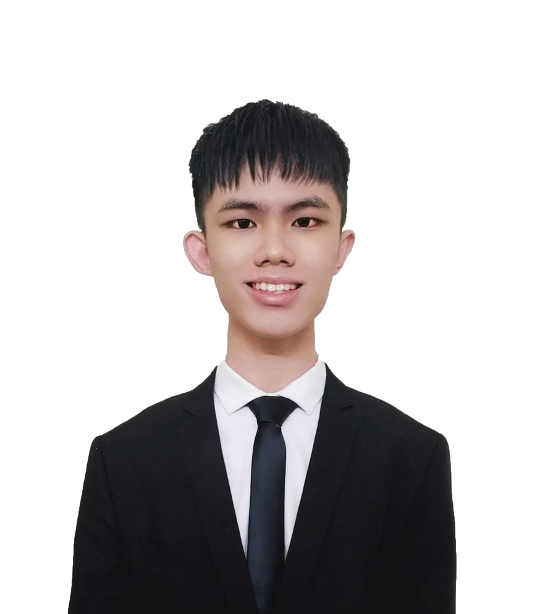}}]{Jing Jie Tan} received his Bachelor of Computer Science (Hons) and PhD in Engineering, specializing in machine learning, from Universiti Tunku Abdul Rahman (UTAR). He is currently a Research Fellow at the National University of Singapore (NUS). His research interests include affective computing, health intelligence, natural language processing, computer vision, deep learning, large language models, and vision transformers. He is passionate about translating research into practical applications, with the goal of leveraging technology to better serve society.
\end{IEEEbiography}

\begin{IEEEbiography}[{\includegraphics[width=1in,height=1.25in,clip,keepaspectratio]{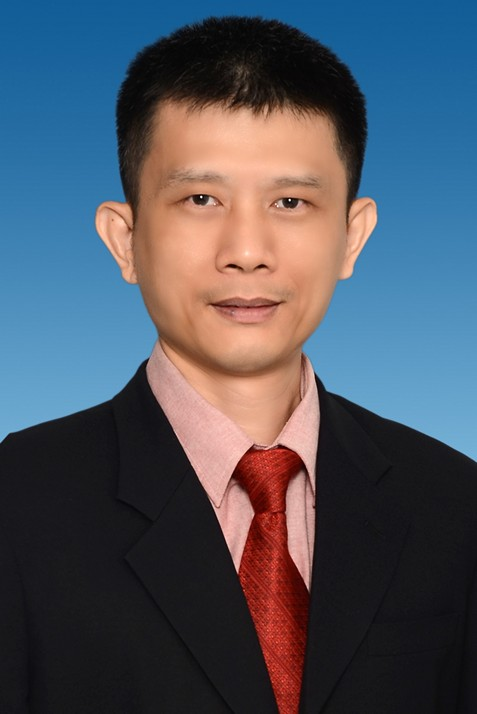}}]{Ban-Hoe Kwan} received the Bachelor of Engineering (Electrical), Master of Engineering Science, and Ph.D. degrees in Engineering from the University of Malaya (UM). He is currently an Associate Professor at Universiti Tunku Abdul Rahman (UTAR). His research interests include image processing, artificial intelligence, medical signal processing, the Internet of Things, and robotics.
\end{IEEEbiography}

\begin{IEEEbiography}[{\includegraphics[width=1in,height=1.25in,clip,keepaspectratio]{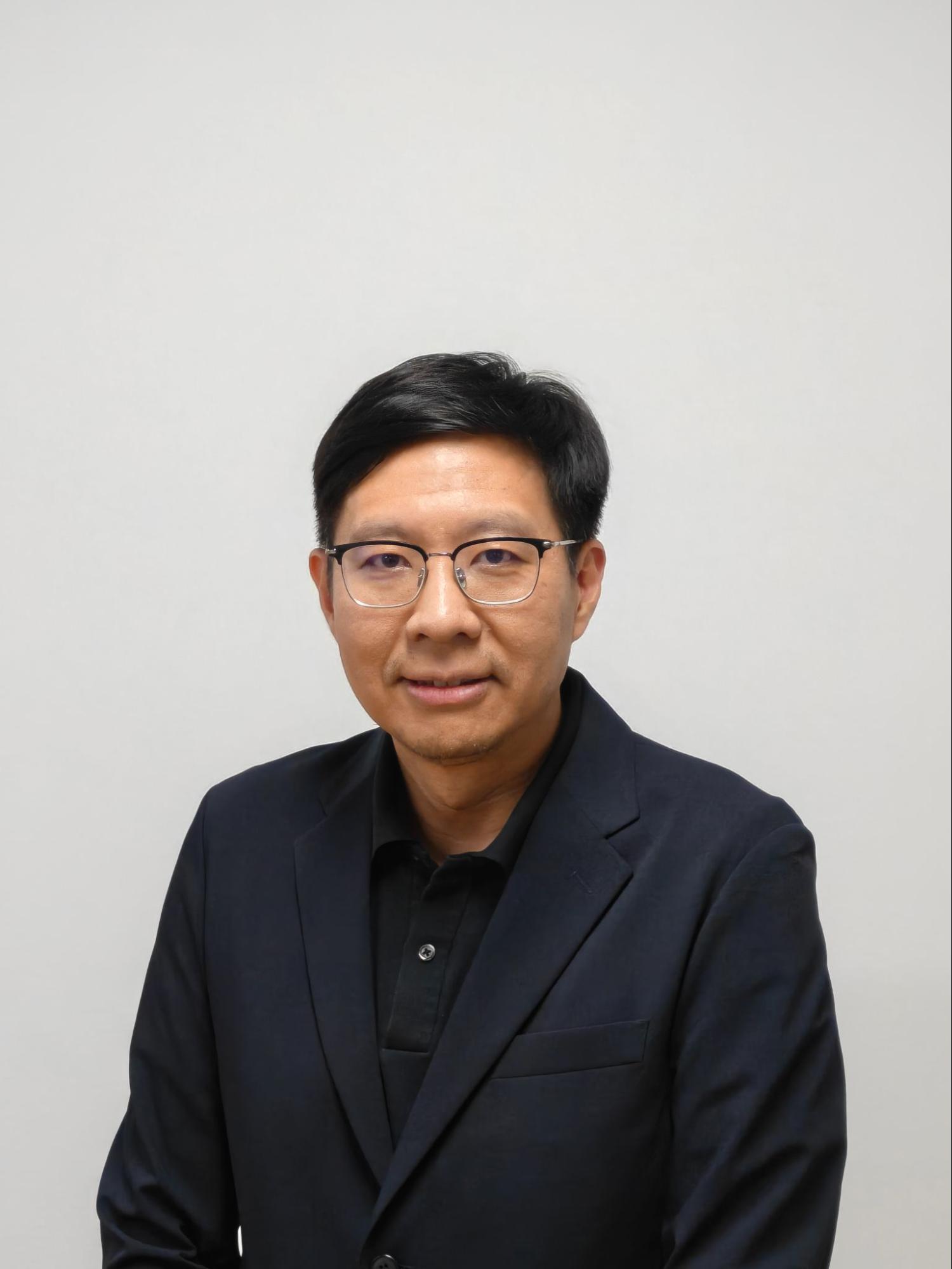}}]{Danny Ng Wee Kiat} (Senior Member, IEEE) received the Ph.D. degree in Engineering and is a registered Professional Engineer with the Board of Engineers Malaysia. He is currently an Assistant Professor at Universiti Tunku Abdul Rahman (UTAR). His research focuses on robotics and artificial intelligence, particularly AI-driven robotics, generative AI, and autonomous AI agents for industrial and enterprise applications. In addition to his academic role, he serves as the Chief Executive Officer and Chief Technology Officer of Netizen Robotics and as the Technical Director of Netizen Experience, where he leads initiatives in advanced robotics and digital transformation.
\end{IEEEbiography}

\begin{IEEEbiography}[{\includegraphics[width=1in,height=1.25in,clip,keepaspectratio]{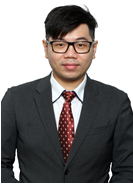}}]{Yan Chai Hum} (Senior Member, IEEE) received the Ph.D. degree in Engineering with specialization in artificial intelligence from Universiti Teknologi Malaysia. He is currently an Associate Professor with the Department of Mechatronics and Biomedical Engineering, Lee Kong Chian Faculty of Engineering and Science, Universiti Tunku Abdul Rahman (UTAR). His research interests include artificial intelligence in healthcare, biomedical imaging, computer vision, Internet of Things systems, and intelligent sensing technologies. His recent work focuses on AI-driven diagnostic systems, multimodal medical data analysis, and autonomous sensing platforms. He is also the founder of Promptiq Enterprise, an AI consultancy specializing in generative AI and intelligent systems.
\end{IEEEbiography}

\begin{IEEEbiography}[{\includegraphics[width=1in,height=1.25in,clip,keepaspectratio]{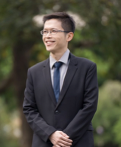}}]{Shih-Yu Lo} is an Associate Professor at the Institute of Communication Studies, National Yang Ming Chiao Tung University, Taiwan. His research integrates cognitive psychology and human–computer interaction, focusing on how emerging technologies such as AI, virtual reality, and social robots shape memory, decision-making, empathy, and social cognition.
\end{IEEEbiography}

\begin{IEEEbiography}[{\includegraphics[width=1in,height=1.25in,clip,keepaspectratio]{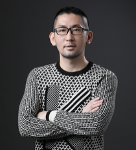}}]{Po-An Chen} is a Professor and Director at the Institute of Information Management, National Yang Ming Chiao Tung University, Taiwan. He is generally interested in economics and computation, artificial intelligence, and operations research, specifically including algorithmic game theory, theoretical online/reinforcement learning, social networks, and multiagent and distributed systems. 
\end{IEEEbiography}

\begin{IEEEbiography}[{\includegraphics[width=1in,height=1.25in,clip,keepaspectratio]{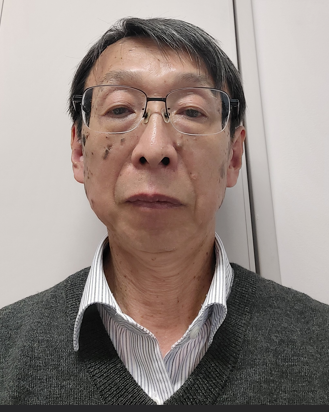}}]{Noriyuki Kawarazaki} received the Ph.D. degree in Engineering from Kyushu University. He is currently a Professor and Department Chair of Information Systems at Kanagawa Institute of Technology, Japan. His research interests include human-robot interaction, image processing, and artificial intelligence in robotics.
\end{IEEEbiography}

\begin{IEEEbiography}[{\includegraphics[width=1in,height=1.25in,clip,keepaspectratio]{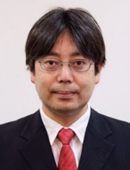}}]{Kosuke Takano} received a B.A. degree in Environment and Information Studies from Keio University, Japan, and his M.A. and Ph.D. degrees in Media and Governance from Keio University. He is currently a Professor in the Department of Information and Computer Sciences at Kanagawa Institute of Technology, Japan. His research interests include emotional AI and multimodal affective computing, data management systems for real-world monitoring, and AI-driven educational systems.
\end{IEEEbiography}

\begin{IEEEbiography}[{\includegraphics[width=1in,height=1.25in,clip,keepaspectratio]{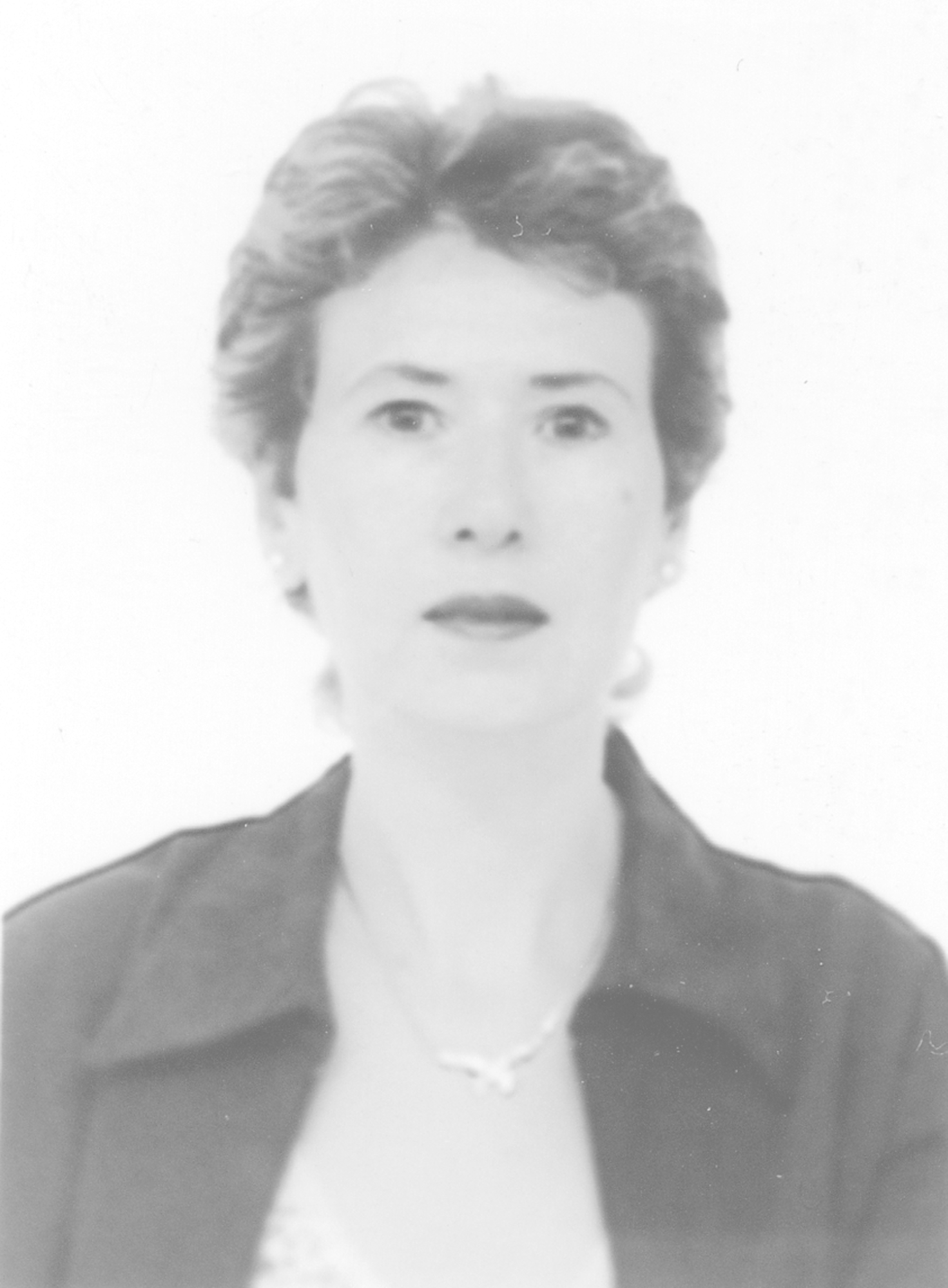}}]{Anissa Mokraoui} received the state engineering degree in electrical engineering from national school of telecommunications in 1989 from Algeria, the M.S degree in information technology in 1990 and the Ph.D. degree in 1994 both from University Paris 11, Orsay France. From 1992 to 1994, she worked at the National Institute of Telecommunications (INT, at Evry France) where her research activities were on digital signal processing, fast filtering algorithms and implementation problems on DSP. In 1997, she was appointed as assistant professor and in December 2011 as associate professor at Galil\'e Institute of University Paris 13, France. Since 2013, she is full professor at Galilée Institute of Universit\'e  Sorbonne Paris Nord (USPN), France. From 2016 to 2024, she was the director of the L2TI laboratory of USPN. Her current research interests include source coding (image, video, multi-view, stereoscopic); joint source-channel-protocol decoding, robust mobile transmission, MIMO-OFDM channel estimation (massive), computer vision, few-shot object detection, cross-domain. She is co-author of more than 150 contributions to journals and conference proceedings. She served on program committees for conferences. She acts as a reviewer for many IEEE and EURASIP conferences and journals.
\end{IEEEbiography}
\end{document}